\documentclass{article} % For LaTeX2e
\usepackage{iclr2027_conference,times}

\usepackage{amsmath,amsfonts,bm}

\def\eqref#1{equation~\ref{#1}}
\def\1{\bm{1}}

\DeclareMathAlphabet{\mathsfit}{\encodingdefault}{\sfdefault}{m}{sl}
\SetMathAlphabet{\mathsfit}{bold}{\encodingdefault}{\sfdefault}{bx}{n}

\usepackage{url}
\usepackage{booktabs}
\usepackage{multirow}
\usepackage{graphicx}
\graphicspath{{figures/}}
\usepackage{amsmath}
\usepackage{xcolor}
\usepackage{bbm}
\usepackage{enumitem}
\definecolor{iris}{rgb}{0.35, 0.31, 0.81}
\definecolor{amaranth}{rgb}{0.9, 0.17, 0.31}
\definecolor{ao}{rgb}{0.0, 0.5, 0.0}
\usepackage[colorlinks=true, linkcolor=amaranth, urlcolor=amaranth, citecolor=iris, anchorcolor=blue]{hyperref}
\usepackage{tabularx}
\usepackage{colortbl}
\usepackage{caption}
\usepackage{array}
\usepackage[capitalize,noabbrev,nameinlink]{cleveref}
\usepackage[skins,breakable]{tcolorbox}
\definecolor{promptblue}{rgb}{0.16, 0.47, 0.84}
\definecolor{promptgreen}{rgb}{0.12, 0.48, 0.31}
\definecolor{promptgrey}{rgb}{0.36, 0.44, 0.47}
\newtcolorbox{promptbox}[2][promptblue]{%
  breakable, enhanced, colback=#1!4, colframe=#1!75!black, coltitle=white,
  fonttitle=\small\bfseries, title={#2}, boxrule=0.7pt, left=5pt, right=5pt,
  top=4pt, bottom=4pt, arc=2pt, before skip=6pt, after skip=6pt}

\title{Humanity's Sixth Sense: Benchmarking Intuitive Visual Reasoning in Multimodal Models}

\author{Xingang Guo$^{1}$\thanks{Equal contribution.} , Jing Gu$^{2*}$, Brian Jang$^{1*}$, Renxiong Wang$^{1}$, Utkarsh Tyagi$^{1}$, Daniel Quigley$^{1}$, \\
\textbf{Steven Li$^{1}$, David Yan$^{2}$, Daniel Yue Zhang$^{1}$, Darvin Yi$^{1}$, Forrest Huang$^{2}$, HiJae Kim$^{1}$,} \\ \textbf{Tianyi Zhang$^{2}$, Jared Lichtarge$^{2}$,  
Jihua Huang$^{2}$, Le Xue$^{2}$, Manan Tomar$^{2}$, Qiuyi Richard Zhang$^{2}$,} \\
\textbf{Ruofei Yu$^{2}$, Seth Neel$^{2}$, Yaning Hu$^{2}$, 
Marcella Valentine$^{2}$, Xinzhe Jiang$^{2}$, Daniel Evans$^{1}$,}\\
\textbf{Chenguang Wang$^{1,3}$, Dustin Tran$^{2}$, Tong Zhao$^{1}$, Yinfei Yang$^{2}$, Yunzhong He$^{1}$}  \\
$^{1}$ScaleAI, $^{2}$Elorian, $^{3}$University of California, Santa Cruz\\
{Leaderboard page: \href{https://scale.com/leaderboard/hss}{https://scale.com/leaderboard/hss}}\\
{Dataset page: \href{https://huggingface.co/datasets/ScaleAI/HSS}{https://huggingface.co/datasets/ScaleAI/HSS}}
}

\iclrfinalcopy  % Uncomment ONLY for the camera-ready; submission stays anonymous.

\begin{document}

\maketitle

\begin{center}
\includegraphics[width=\textwidth]{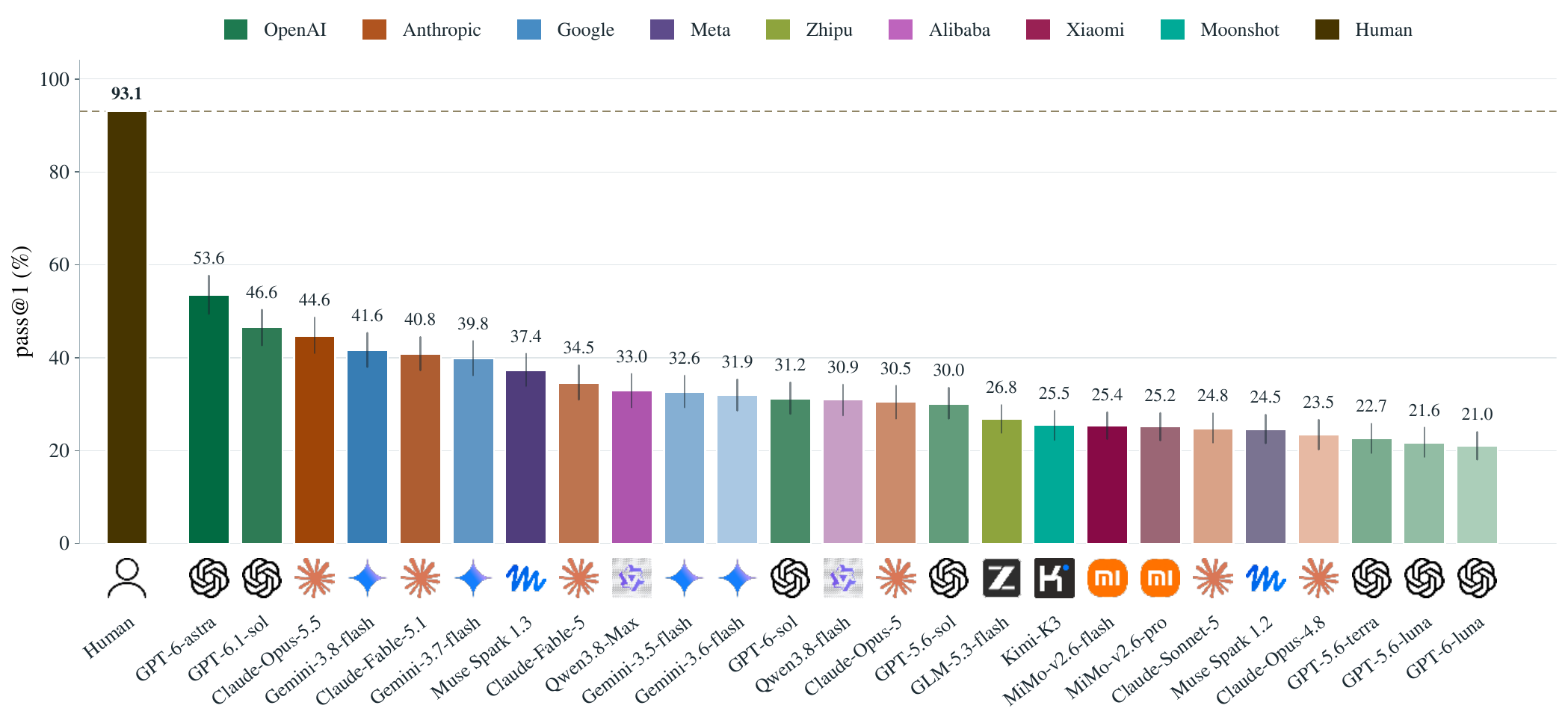}
\captionof{figure}{\textbf{HSS leaderboard.} Performance on HSS for $25$
multimodal models. Every model scores far below people: the best, GPT-6-astra, reaches $53.6\%$ where humans reach $93.1\%$.}
\label{fig:leaderboard}
\end{center}

\begin{abstract}
Humans perceive far more in a scene than what is explicitly depicted: a single glance captures past causes and future trajectories; a quick peek determines if a vehicle can fit between two parked cars; a few seconds of video reveals who holds authority in a room; and a fleeting clip highlights subtle abstract patterns like unwritten rules or hidden labels. This capacity reflects a form of humanity's \textit{sixth sense}: an intuitive reasoning mechanism that recovers implicit information beyond raw sensory perception. Crucially, this rapid, zero-shot visual intuition underpins everyday navigation and social interaction, making it a vital capability for Multimodal Large Language Models (MLLMs) deployed alongside people. Existing visual benchmarks, however, target either deliberate expert-level analysis in academic and mathematical domains or low-level perception, leaving the intuitive reasoning that people perform largely untested. To bridge this gap, we introduce \textbf{Humanity's Sixth Sense (HSS)}, a benchmark for intuitive visual reasoning. HSS spans diverse image and video inputs, organizes items under a structured taxonomy, and pairs each with human-written prompts probing the implicit temporal, spatial, social, and abstract structure that people infer at a glance. Frontier MLLMs fall short of human performance: participants reach 93.1\% accuracy, while the strongest model, GPT-6-astra, reaches only 53.6\% even at maximum reasoning effort. Despite excelling in many complex tasks that require advanced perception and knowledge, current models still struggle significantly on these visual tasks that are intuitive for humans. We further explore agentic setup that apply dynamic visual manipulation to HSS, which narrows but does not close the gap. HSS establishes intuitive visual reasoning as a measurable axis and directs attention to a capability that scaling on current benchmarks has so far left behind.
\end{abstract}

% ===========================================================================
\section{Introduction}
\label{sec:introduction}
% ===========================================================================

% \looseness=-1
A single glance at a visual scene reveals far more than what is explicitly rendered in pixels. Within a fraction of a second, we intuitively register who holds authority in a room, sense the subtle tension behind a polite smile, or notice a toddler edging toward the stairs, long before anything goes wrong. We call this capacity humanity’s \emph{sixth sense}: a rapid, pre-verbal form of visual inference that transforms circumstantial evidence into actionable conclusions. For AI agents deployed in homes, vehicles, and workplaces, this skill is not a luxury, but a core necessity: an agent must \emph{perceive the world through a human lens} so that its predictions and behaviors naturally align with the people around it. This capacity relies on \emph{visual intuition}. In cognitive science, intuition is characterized as an experience-based process that yields spontaneous judgments without conscious derivation \citep{kotler2025pathfinding}, functioning as skilled pattern recognition in environments with learnable regularities \citep{kahneman2009conditions}. In human vision, this is remarkably concrete: people categorize natural scenes within roughly $150$ms \citep{thorpe1996speed} while effortlessly extracting implied physical dynamics, human motives, and social structures \citep{spelke2007core,battaglia2013simulation,baker2017rational}. We synthesize these insights to define an inference as \emph{intuitive} when it is fast, automatic, grounded in everyday experience rather than step-by-step derivation, and accessible even when its underlying rationale is difficult to articulate. 

In text-only domains, benchmarks like SimpleBench \citep{philip2024simplebench} have shown that frontier models lag significantly behind humans on commonsense spatio-temporal and social reasoning. In vision, the closest counterpart is BabyVision \citep{chen2026babyvision}, which highlights model failures on basic visual tasks that young children solve effortlessly. However, while BabyVision tests whether models perceive what is \emph{explicitly present}, HSS evaluates whether they infer what is \emph{implicit}. Visual scenes carry vast unstated context such as revealing interpersonal relationships, latent intentions, physical affordances, and imminent consequences without ever stating them. Existing visual benchmarks fail to probe this dimension: they either treat images as prompts for specialized academic problem-solving (e.g., MMMU \citep{yue2024mmmu}, HLE \citep{phan2025hle}) or isolate narrow perceptual capabilities like visual searching and object tracking (e.g., V* \citep{wu2024vstar}, Video-MME \citep{fu2024videomme}, Video-MME-v2 \citep{fu2026video}). Meanwhile, existing probes of physical and social intelligence \citep{zellers2019vcr,riochet2021intphys,kamath2023whatsup,zadeh2019socialiq,bear2021physion,jin2024mmtomqa} remain fragmented across isolated domains, rely heavily on synthetic environments, or depend on multiple-choice formats that allow models to succeed through process-of-elimination heuristics rather than true visual inference.

\begin{figure}[t]
\centering
\includegraphics[width=\textwidth]{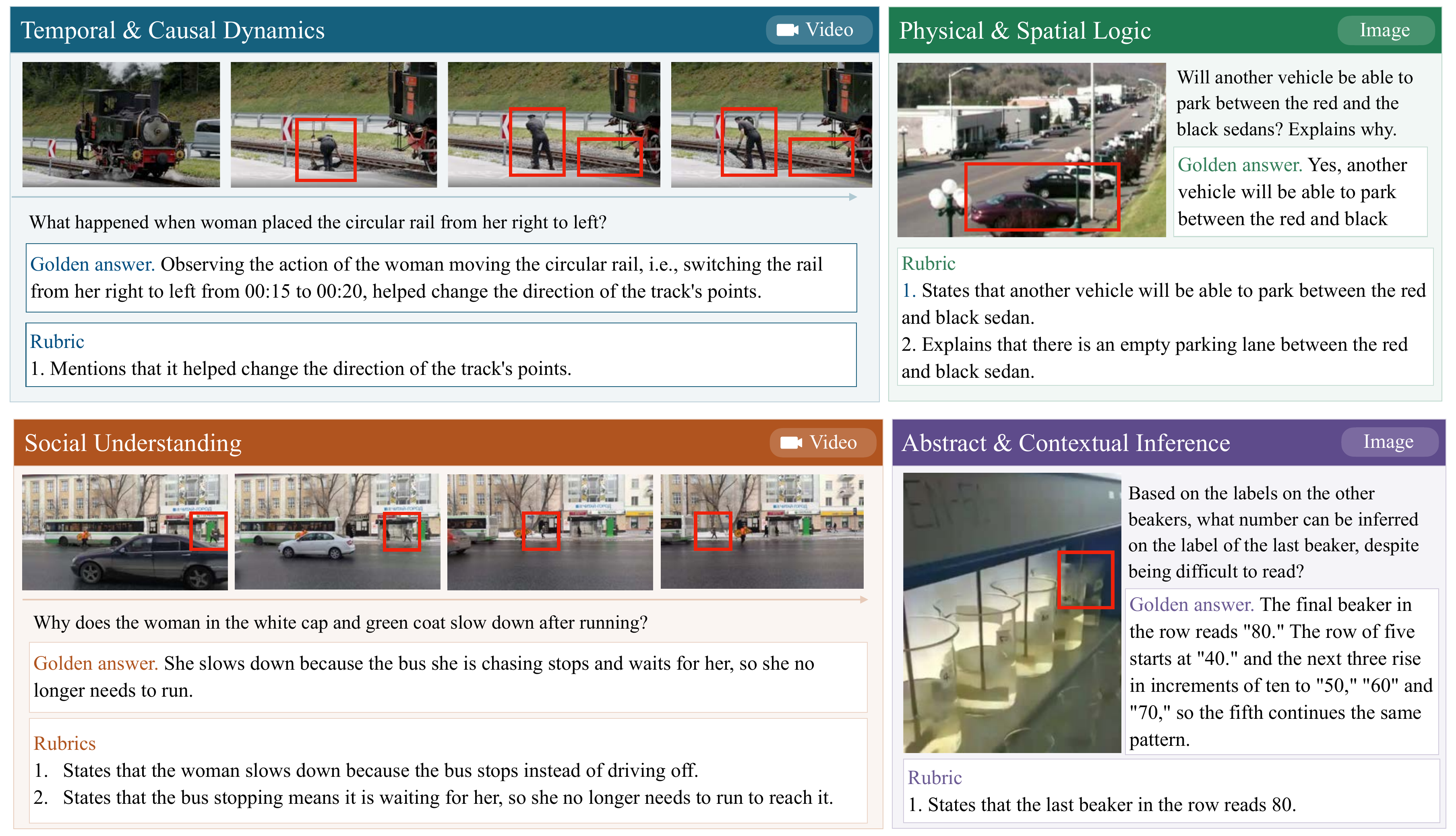}
\caption{\textbf{HSS task examples across the four domains.} Each panel shows the input media (a single image or sampled video keyframes), human-written question, reference answer, and evaluation rubric. \emph{Temporal \& causal dynamics}: identifying the physical outcome of a manual track-switching action in video. \emph{Physical \& spatial logic}: judging whether a vehicle can fit between two parked sedans without explicit scale markers. \emph{Social understanding}: inferring why a person stops running after a bus in a video clip. \emph{Abstract \& contextual inference}: deducing an occluded beaker label from an arithmetic progression of neighboring legible labels.} % Appendix~\ref{app:domain_media_examples} shows two further tasks for each domain.}
\label{fig:examples_main}
\vspace{-0.1in}
\end{figure}

To address this gap, we introduce \textbf{Humanity’s Sixth Sense (HSS)}, a benchmark specifically targeting the domain of intuitive visual reasoning that operates between raw perception and deliberate, step-by-step logic. Concretely, an HSS task requires interpreting visual evidence alongside contextual, physical, social, and commonsense knowledge to deduce unstated implications, latent states, and likely outcomes. For instance, viewers must infer an agent's intended goal, recognize physical affordances, reconstruct prior events, or anticipate imminent consequences purely from visual cues and everyday experience. HSS comprises open-ended tasks spanning four core domains: \emph{temporal \& causal dynamics}, \emph{physical \& spatial logic}, \emph{social understanding}, and \emph{abstract contextual inference} (Figure~\ref{fig:examples_main}), categorized into 11 subdomains that reflect human scene interpretation (Section~\ref{sec:taxonomy}). To avoid multiple-choice artifacts, models generate free-form responses evaluated against human-annotated atomic rubrics. 

% We rigorously validate this pipeline, demonstrating that human reference answers achieve 100\% accuracy and that diverse judge models exhibit over 95\% evaluation agreement.

Our experiments reveal a substantial capability gap (Figure~\ref{fig:leaderboard}). Human participants achieve $93.1\%$ pass rate on HSS, whereas the strongest model evaluated, GPT-6-astra, attains only $53.6\%$ even at maximum reasoning effort. Moreover, standard scaling strategies do not close this gap: increased test-time deliberation, higher reasoning effort, and agentic tool use each yield only modest improvements, and performance remains far from saturation. These results indicate that intuitive visual reasoning constitutes a distinct and largely unexplored dimension of visual intelligence, one that does not emerge simply from scaling test-time compute.

Our contributions are summarized as follows:
\begin{enumerate}[leftmargin=1.5em,itemsep=2pt,topsep=2pt]
\item \textbf{A structured taxonomy of intuitive visual reasoning.} We formalize intuitive visual reasoning across four core domains and eleven subdomains, establishing a principled cognitive framework for evaluating implicit visual inference.
\item \textbf{A rigorous, human-verified benchmark.} We introduce HSS, a benchmark specifically curated to assess implicit visual reasoning. To address the inherent subjectivity of human intuition, all instances are authored by human annotators and pass through a multi-stage independent review pipeline to guarantee strict human consensus and high data precision.
\item \textbf{Comprehensive empirical evaluation.} Evaluating state-of-the-art MLLMs under direct-inference reveals that current MLLMs achieve barely half of human performance ($53.6\%$ vs.\ $93.1\%$), establishing HSS as a formidable open challenge.
\item \textbf{Agentic paradigms and ablation analyses.} We examine the effects of agentic tool-use workflows and reasoning effort, demonstrating that intuitive visual reasoning represents a capability gap that remains resistant to both agentic augmentation and standard compute scaling.
\end{enumerate}

% ===========================================================================
\section{The HSS Benchmark}
\label{sec:benchmark}
% ===========================================================================

\subsection{Task Format}
\label{sec:task_format}

Each benchmark instance is a tuple $(m, q, g, R)$: a visual media $m$ (a static image
or a video clip), a natural-language question $q$, a human-authored reference answer
$g$, and a set of atomic rubrics $R = \{r_1, \dots, r_k\}$,
each independently checkable against $g$ and \textbf{mandatory} for an answer to count as correct. Given $m$ and $q$, a model returns a free-form
answer $a$, and a judge marks every criterion $r_i \in R$ as met or unmet given $q$, $a$,
$g$, and $R$; an instance counts as solved only when all $k$ are met
(Section \ref{sec:metrics}).

\subsection{HSS Taxonomy Design}
\label{sec:taxonomy}

Human intuitive visual reasoning enables us to infer unobserved dynamics, physical constraints, mental states, and contextual rules directly from visual input. Our taxonomy builds on three core domains established in cognitive science, including core knowledge of objects and space \citep{spelke2007core}, intuitive physics \citep{battaglia2013simulation}, and mentalizing \citep{baker2017rational}, and further introducing an additional domain for abstract regularities implied by a scene. To render this latent capacity measurable, we formalize it into four core domains encompassing eleven subdomains (Table~\ref{tab:hss-taxonomy}):

\begin{enumerate}[leftmargin=1.5em,itemsep=2pt,topsep=2pt]
    \item \textbf{Temporal \& Causal Dynamics (Time):} Reconstructs the unobserved timeline connecting visual states, evaluating inferences about past events (\textit{Retrodiction}), underlying physical drivers (\textit{Mechanistic Causality}), and future trajectories (\textit{Extrapolation}).
    \item \textbf{Physical \& Spatial Logic (Space):} Reasons over unshown content (\textit{Hidden \& Invisible}), functional interactions (\textit{Affordance \& Feasibility}), novel perspectives (\textit{Spatial ``Alien Viewpoint''}), and physical accessibility under spatial constraints (\textit{Spatial Reachability}).
    \item \textbf{Social Understanding (Mind):} Formalizes the invisible mental and normative substrate of human interaction, capturing intent and belief attribution (\textit{Theory of Mind}) as well as implicit social expectations (\textit{Social Role, Norm \& Power Dynamics}).
    \item \textbf{Abstract \& Contextual Inference (Context):} Derives non-perceptual implications that a scene implies but never explicitly depicts, assessing unseen state transitions (\textit{Change \& Consequence}) and structural regularities (\textit{Patterns \& Pareidolia}).
\end{enumerate}

Rather than treating visual intuition as an unanalyzable or monolithic capacity, the taxonomy described above provides a principled foundation that fulfills a dual purpose: first, as a generative blueprint that enforces rigorous, multi-faceted coverage during benchmark curation; and second, as a fine-grained diagnostic framework for model evaluation. By projecting performance across these eleven cognitive dimensions rather than relying solely on aggregate scores, the taxonomy disentangles distinct competency profiles and exposes the specific latent reasoning gaps of state-of-the-art multimodal models.

\subsection{Task Design Principles}
\label{sec:design_principles}

Because intuitive visual reasoning inherently risks subjective interpretation, task curation in HSS is strictly governed by three core design principles:

\begin{enumerate}[leftmargin=1.5em,itemsep=3pt,topsep=2pt]
    \item \textbf{Exclusion of Pure Perception:} Tasks must require latent inference beyond what is explicitly depicted in the visual media. We reject tasks that rely purely on fine-grained visual recognition, regardless of visual complexity. If a high-resolution crop of the target region makes the answer immediately obvious, the task evaluates visual acuity rather than intuitive reasoning and is discarded.
    \item \textbf{Exclusion of Specialized Expertise:} Tasks must target intuitive reasoning accessible to any attentive adult. We strictly prohibit items requiring symbolic calculation, specialized academic knowledge, or niche professional training, ensuring the benchmark purely isolates common-sense visual intuition.
    \item \textbf{Strict Human Consensus and Unambiguity:} Both questions and target answers must command unambiguous human consensus. Prompts are carefully structured to prevent subjective or divergent interpretations, and a task is admitted into the benchmark only if independent reviewers reach unanimous agreement on the ground truth.
\end{enumerate}

\subsection{Benchmark Construction Pipeline}
\label{sec:construction}

Each task in HSS is authored by a trained human annotator through a three-step workflow: (1) selecting a source image or video clip from a curated media pool; (2) assigning the item to one of our four core domains and eleven subdomains; and (3) composing a natural-language prompt $q$, a gold reference answer $g$, and a structured rubric $R$ decomposing $g$ into fine-grained, independently checkable criteria. Authored tasks then enter a rigorous quality-control pipeline consisting of three independent review rounds. At each stage, a reviewer renders one of three verdicts: \textbf{Pass:} Advances the task to the subsequent review stage. \textbf{Edit:} Returns the task to the authoring phase for revision with specific feedback, after which it must re-enter the review pipeline from the beginning. \textbf{Drop:} Permanently removes the item if the prompt is ambiguous, the visual media is uninformative, or the target inference fails to command unanimous human consensus.

Only tasks that achieve unanimous approval across all three independent rounds are admitted into the benchmark, guaranteeing that every released rubric has been verified by at least three reviewers beyond its author. The review pipeline is highly lossy, and this attrition is the primary driver of dataset quality: out of $3{,}466$ initially authored tasks, $1{,}164$ survive the first review stage, $559$ pass the second, and $522$ pass the third, yielding an overall end-to-end acceptance rate of $15.1\%$. The most significant filter occurs at the first review gate, where approximately two-thirds of submitted tasks are filtered out, primarily for relying on pure visual perception or exhibiting subjective prompts and target answers. Appendix~\ref{app:HSS} (Figure~\ref{fig:pipeline}) details the complete flowchart and attrition mechanics across each review stage. The final released benchmark comprises these $522$ strictly vetted tasks across 4 domains and 11 subdomains ($288$ image-based and $234$ video-based instances), carrying $723$ fine-grained rubric criteria in total (Table~\ref{tab:dataset}).

% ===========================================================================
\section{Experiments}
\label{sec:results}
% ===========================================================================

We evaluate 25 representative MLLMs on HSS. For models supporting native video inputs, we feed raw video files directly. Otherwise, we apply uniform frame sampling capped at 500 frames; for models with stricter input limits, we default to their maximum supported frame capacity. Static images are evaluated at native resolution, downscaling only when required by API or payload constraints. Model outputs are graded by Claude-Opus-5, serving as an automated LLM judge. Because the HSS data curation process retains only mandatory criteria and excludes non-essential ones, a response is marked as successful if and only if it satisfies all of the corresponding rubric criteria. Further details regarding the experimental setup, prompts, and evaluation protocol are in Appendix~\ref{app:exp_setup}.

\subsection{Benchmark Results}

Table \ref{tab:main} reports $25$ models from eight vendors on HSS. Several key findings stand out:

\textbf{HSS is far from saturated despite high human performance.} The top-performing model, GPT-6-astra (maximum reasoning effort), achieves only a $53.6\%$ pass rate, compared to $93.1\%$ for human annotators. Most models perform below $40\%$. In addition, across all $25$ models, the average reasoning token usage is around $4{,}000$ per task. Even on tasks intuitive to humans, models generate substantial reasoning chains before answering. While extended chain-of-thought reflects current reasoning alignment techniques, models frequently overthink without arriving at the correct answer for HSS.

\textbf{Results breakdown across media type and domains.} Video tasks prove more challenging than static images for $23$ of the $25$ models, with an average drop of $7.3$ percentage points. This gap highlights ongoing struggles with long temporal horizons and key-frame localization. Social understanding is the weakest domain across models; it is the lowest-performing domain for $21$ of the $25$ models, averaging $24.4\%$ accuracy compared to $34.1\%$ across the remaining three domains.

\textbf{Visual inputs are indispensable, preventing pure language-prior guessing.} The blind control experiment confirms that HSS cannot be solved through language priors alone: stripping visual media drops the top model's score from $53.6\%$ to $6.6\%$. The residue is guessing rather than knowledge: only $4$ of $522$ tasks pass blind on all three attempts, and the passes concentrate on two-way questions, where the model commits to one side and invents supporting detail (Appendix~\ref{app:blind_test_example}).

\textbf{Human performance.} Twenty annotators answered the tasks under the same
free-form conditions as the models and reached $93.1\%$, ahead of the best model by
nearly $40$ points. Each answer also carries a self-reported confidence. Annotators were
unsure on only $63$ of $522$ tasks ($12.1\%$) and scored $94.1\%$ on the $459$ they were sure about. Even the unsure answers scored $85.7\%$: the competence HSS measures stays available to people when their confidence is not.

\begin{table}[t]
\caption{\textbf{Main results.} Pass@1 (\%) on HSS. A task counts as correct only when every rubric criterion is met; $\pm$ is the half-width of a $95\%$ bootstrap interval over tasks. The columns re-cut the same tasks by media type and by domain, and \emph{think-tok} / \emph{ans-tok} are mean thinking and answer tokens per task. The best model value is \textbf{bold} and the second best is \underline{underlined}, and the \emph{blind} control re-runs the strongest model with the media removed.}
\label{tab:main}
\centering
\resizebox{1.0\linewidth}{!}{
\begin{tabular}{ll|c|cc|cccc|cc}
\toprule
\textbf{Vendor} & \textbf{Model (Reasoning Effort)}  & pass@1 & image & video & temporal & physical & social & abstract & think-tok & ans-tok\\
\midrule
 & Human baseline & $\mathbf{93.1}$ & $94.1$ & $91.9$ & $97.0$ & $94.9$ & $89.1$ & $89.2$ & --- & --- \\
 \midrule
\multirow{7}{*}{OpenAI} & GPT-6-astra (max) & $\mathbf{53.6}$\,\tiny$\pm4.1$ & $\mathbf{55.6}$ & $\mathbf{51.1}$ & $\mathbf{58.2}$ & $\mathbf{53.6}$ & $\mathbf{46.1}$ & $\mathbf{55.6}$ & $2{,}896$ & $40$ \\
 & GPT-6.1-sol (max) & $\underline{46.6}$\,\tiny$\pm3.8$ & $45.5$ & $\underline{47.9}$ & $\underline{55.2}$ & $\underline{45.8}$ & $36.4$ & $47.4$ & $2{,}298$ & $36$ \\
 & GPT-6-sol (max) & $31.2$\,\tiny$\pm3.4$ & $31.8$ & $30.5$ & $38.1$ & $28.6$ & $22.4$ & $36.3$ & $3{,}298$ & $30$ \\
 & GPT-5.6-sol (max) & $30.0$\,\tiny$\pm3.3$ & $28.5$ & $31.9$ & $38.6$ & $26.3$ & $27.3$ & $28.1$ & $3{,}630$ & $39$ \\
 & GPT-5.6-terra (max) & $22.7$\,\tiny$\pm3.2$ & $24.1$ & $20.9$ & $28.4$ & $21.0$ & $19.1$ & $21.9$ & $5{,}794$ & $45$ \\
 & GPT-5.6-luna (max) & $21.6$\,\tiny$\pm3.2$ & $23.5$ & $19.4$ & $26.9$ & $24.6$ & $13.3$ & $18.6$ & $5{,}146$ & $42$ \\
 & GPT-6-luna (max) & $21.0$\,\tiny$\pm3.0$ & $23.3$ & $18.2$ & $26.9$ & $24.1$ & $9.4$ & $20.6$ & $3{,}993$ & $39$ \\
\midrule
\multirow{6}{*}{Anthropic} & Claude-Opus-5.5 (xhigh) & $44.6$\,\tiny$\pm3.8$ & $\underline{51.3}$ & $36.5$ & $44.0$ & $44.9$ & $34.8$ & $\mathbf{55.6}$ & $1{,}496$ & $294$ \\
 & Claude-Fable-5.1 (xhigh) & $40.8$\,\tiny$\pm3.6$ & $44.7$ & $36.0$ & $43.3$ & $41.1$ & $34.2$ & $44.1$ & $3{,}478$ & $360$ \\
 & Claude-Fable-5 (max) & $34.5$\,\tiny$\pm3.7$ & $37.2$ & $31.3$ & $38.1$ & $37.1$ & $30.0$ & $30.4$ & $3{,}880$ & $463$ \\
 & Claude-Opus-5 (max) & $30.5$\,\tiny$\pm3.5$ & $32.5$ & $27.9$ & $31.3$ & $35.0$ & $22.4$ & $30.1$ & $2{,}888$ & $445$ \\
 & Claude-Sonnet-5 (max) & $24.8$\,\tiny$\pm3.2$ & $29.5$ & $18.9$ & $29.1$ & $27.5$ & $14.2$ & $25.8$ & $3{,}209$ & $408$ \\
 & Claude-Opus-4.8 (max) & $23.5$\,\tiny$\pm3.2$ & $27.2$ & $18.9$ & $27.1$ & $26.5$ & $14.2$ & $23.5$ & $2{,}416$ & $408$ \\
\midrule
\multirow{4}{*}{Google} & Gemini-3.8-flash (high) & $41.6$\,\tiny$\pm3.7$ & $47.2$ & $34.8$ & $43.3$ & $37.5$ & $\underline{40.0}$ & $\underline{48.4}$ & $10{,}407$ & $158$ \\
 & Gemini-3.7-flash (high) & $39.8$\,\tiny$\pm3.7$ & $45.4$ & $33.0$ & $40.5$ & $40.9$ & $35.8$ & $41.5$ & $1{,}858$ & $118$ \\
 & Gemini-3.5-flash (high) & $32.6$\,\tiny$\pm3.4$ & $35.1$ & $29.6$ & $35.6$ & $31.6$ & $31.8$ & $31.4$ & $2{,}456$ & $184$ \\
 & Gemini-3.6-flash (high) & $31.9$\,\tiny$\pm3.4$ & $32.5$ & $31.2$ & $37.1$ & $29.7$ & $32.4$ & $28.4$ & $3{,}166$ & $178$ \\
\midrule
\multirow{2}{*}{Meta} & Muse Spark 1.3 (max) & $37.4$\,\tiny$\pm3.5$ & $44.8$ & $28.2$ & $39.6$ & $38.1$ & $25.8$ & $45.8$ & $6{,}278$ & $281$ \\
 & Muse Spark 1.2 (xhigh) & $24.5$\,\tiny$\pm3.1$ & $27.5$ & $20.8$ & $23.6$ & $24.2$ & $21.2$ & $29.7$ & $3{,}640$ & $412$ \\
\midrule
\multirow{2}{*}{Alibaba} & Qwen3.8-Max (max) & $33.0$\,\tiny$\pm3.6$ & $38.3$ & $26.2$ & $38.1$ & $35.2$ & $21.0$ & $34.7$ & $7{,}423$ & $217$ \\
 & Qwen3.8-flash (max) & $30.9$\,\tiny$\pm3.3$ & $37.2$ & $22.9$ & $37.9$ & $29.2$ & $18.6$ & $37.1$ & $7{,}260$ & $200$ \\
\midrule
\multirow{1}{*}{Zhipu} & GLM-5.3-flash (max) & $26.8$\,\tiny$\pm3.0$ & $30.2$ & $22.6$ & $31.3$ & $26.3$ & $18.2$ & $31.0$ & $4{,}484$ & $247$ \\
\midrule
\multirow{1}{*}{Moonshot} & Kimi-K3 (max) & $25.5$\,\tiny$\pm3.1$ & $29.1$ & $21.2$ & $33.1$ & $27.4$ & $15.5$ & $23.2$ & $4{,}949$ & $98$ \\
\midrule
\multirow{2}{*}{Xiaomi} & MiMo-v2.6-flash (high) & $25.4$\,\tiny$\pm2.9$ & $29.2$ & $20.6$ & $30.1$ & $29.6$ & $12.1$ & $26.7$ & $2{,}294$ & $182$ \\
 & MiMo-v2.6-pro (high) & $25.2$\,\tiny$\pm2.9$ & $30.4$ & $18.6$ & $28.3$ & $29.5$ & $14.5$ & $25.7$ & $2{,}499$ & $192$ \\
\midrule
\multirow{1}{*}{Control} & GPT-6-astra (max) [blind] & $6.6$\,\tiny$\pm1.5$ & $2.4$ & $11.7$ & $9.5$ & $5.5$ & $4.8$ & $6.5$ & $1{,}046$ & $41$ \\
\bottomrule
\end{tabular}}
\end{table}

\subsection{Model Failure Analysis}
\label{sec:failures}
To investigate why models fail, we categorize $8{,}573$ failed first attempts from the main evaluation using an eleven-cause taxonomy grouped into three families. \textbf{Perception}: errors related to properties explicitly present in the visual signal (e.g., scale, geometry, facing direction, and spatial alignment). Extending beyond low-level object detection (e.g., recognizing a person or car), this involves evaluating visible visual evidence for intuitive reasoning. \textbf{Latent Inference}: errors related to information not directly rendered in pixels (e.g., unobserved past events, hidden quantities behind obstacles, or underlying motives), requiring the model to bridge the gap between what is seen and what remains unobserved or implied. \textbf{Reasoning}: errors related to logical deduction, where the model fails to synthesize correct explicit and latent visual facts to draw valid conclusions.

\begin{figure}[t]
\centering
\includegraphics[width=\textwidth]{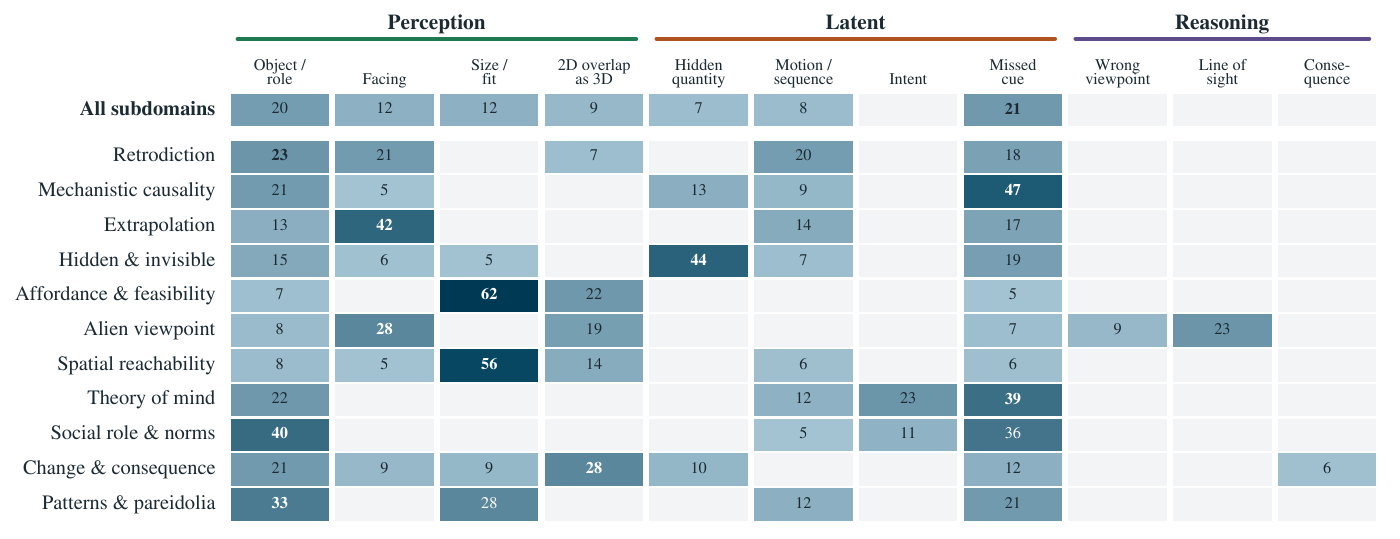}
\caption{\textbf{What models get wrong, by subdomain.} Share (\%) of each subdomain's
failures attributed to each cause, over $8{,}573$ labelled failures from $24$ models on
$522$ tasks.}
\label{fig:failure_causes}
\vspace{-0.2in}
\end{figure}

As shown in Figure~\ref{fig:failure_causes}, failures stem predominantly from visual perception ($53\%$) and latent inference ($41\%$), whereas abstract reasoning accounts for only $5\%$. Overall, the two largest single failure modes are missing a decisive visual cue ($21\%$) and misidentifying an object, person, or role ($20\%$). Furthermore, failure patterns are highly systematic across architectures rather than model-specific: on tasks where five or more models fail, a median of $80\%$ fail due to the exact same cause. Subdomains also exhibit distinct failure signatures: \emph{affordance} and \emph{spatial reachability} fail primarily on size and fit ($62\%$ and $56\%$), \emph{extrapolation} on facing direction ($42\%$), \emph{hidden \& invisible} on hidden quantities ($44\%$), and \emph{mechanistic causality} and \emph{theory of mind} on missed cues ($47\%$ and $39\%$). \emph{Alien viewpoint} is the only subdomain with a notable reasoning component, led by line-of-sight errors ($23\%$). Extended failure analyses and case studies are provided in Appendices~\ref{app:failures} and \ref{app:subdomain_cases}.

\subsection{Agentic Harnesses}
\label{sec:agentic}
\begin{figure}[t]
\centering
\includegraphics[width=\textwidth]{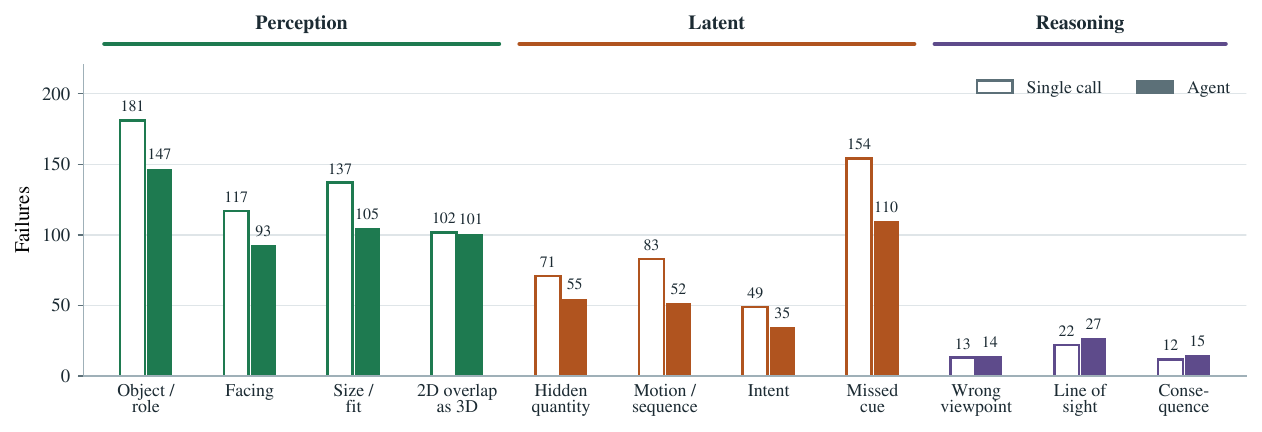}
\caption{\textbf{What the agentic scaffold fixes.} Pooled over the four harness/base
pairs of Table~\ref{tab:agentic_full}, first attempts. The agent cuts failures from
$944$ to $759$, mostly on causes a closer look can fix: object/role
($181 \rightarrow 147$), facing ($117 \rightarrow 93$), size/fit
($137 \rightarrow 105$), and missed-cue ($154 \rightarrow 110$) errors.
Depth errors (a 2D overlap read as 3D alignment) do not fall
($102 \rightarrow 101$), and the small reasoning causes grow.}
\label{fig:agentic_failures}
\vspace{-0.1in}
\end{figure}

To test whether active tool use aids performance on HSS, we evaluate base models inside vendor-provided agentic harnesses, including Claude Code and Codex across $388$ tasks (a subset of HSS). As shown in Table~\ref{tab:agentic}, the agentic setup improves accuracy across all configurations, though performance gains and compute overhead vary widely. Claude Code provides substantial accuracy increases ($+16.5$ to $+20.5$ points) at high compute costs (median $57$--$72$ steps; $>42\text{k}$ thinking tokens per task), whereas Codex yields more modest improvements ($+4.9$ to $+7.5$ points) with minimal overhead ($14$--$18$ steps). However, even the top-performing setup (Codex with GPT-6-astra) reaches only $59.3\%$, demonstrating that HSS remains far from saturated even when models are equipped with agentic capabilities. Re-labeling agentic failures using our error taxonomy reveals distinct patterns in where agent loops succeed and fail:
\begin{itemize}[leftmargin=1.5em,itemsep=2pt,topsep=2pt]
\item \textbf{Active re-inspection repairs perception, not spatial depth:} Total failures decrease from $944$ to $759$ (Figure~\ref{fig:agentic_failures}), driven by a $19$--$37\%$ reduction in perception and latent errors (e.g., recovering missed cues or misidentified objects). However, 2D-to-3D projection errors persist, demonstrating that zooming into a 2D image cannot recover lost 3D spatial depth.
\item \textbf{Gains concentrate in evidence-rich subdomains:} Subdomains where visual evidence is explicitly present in-frame benefit most, such as \emph{patterns \& pareidolia} ($+30.6$ points) and \emph{retrodiction} ($+18.9$ points; Table~\ref{tab:agentic_subdomain}). Tasks whose answer turns on depth or on an unobserved outcome gain least (e.g., \emph{spatial reachability} at $+4.8$ points).
\item \textbf{Agentic failure dynamics:} Agent loops introduce unique failure modes, occasionally over-analyzing inputs and talking themselves out of correct initial readings. Additionally, failed agent runs take significantly more steps than successful ones (median $81$ vs. $67$ steps).
\end{itemize}
Appendix~\ref{app:agentic} reports all four harness/base pairs and provides four annotated trajectories; Appendix~\ref{app:failures} details the complete failure analysis for agentic experiments.

\begin{table}[ht]
\caption{\textbf{Agentic harnesses against single-pass inference}. \emph{Raw} is the base model called once, \emph{Agent} the same model inside
the harness; every $\Delta$ is agentic minus single-pass on those tasks, re-cut by media
type and domain as in Table~\ref{tab:main}. \emph{think-tok} is mean thinking tokens per
task inside the harness and \emph{steps} the median number of agent steps.}
\label{tab:agentic}
\centering
\resizebox{1.0\linewidth}{!}{
\begin{tabular}{ll|ccc|cc|cccc|rr}
\toprule
 & & \multicolumn{3}{c|}{\textbf{Overall pass@1}} & \multicolumn{2}{c|}{\textbf{$\Delta$ by media}} & \multicolumn{4}{c|}{\textbf{$\Delta$ by domain}} & \multicolumn{2}{c}{\textbf{Cost}} \\
\textbf{Harness} & \textbf{Base model} & raw & agent & $\Delta$ & image & video &
temporal & physical & social & abstract & think-tok & steps \\
\midrule
\multirow{2}{*}{Claude Code} & Claude-Opus-5 & $30.8$ & $51.3$ & $\mathbf{+20.5}$ & $\mathbf{+24.3}$ & $+15.8$ & $\mathbf{+26.3}$ & $\mathbf{+15.1}$ & $+18.2$ & $\mathbf{+25.6}$ & $43{,}764$ & $72$ \\
 & Claude-Fable-5.1 & $41.2$ & $57.7$ & $+16.5$ & $+15.4$ & $\mathbf{+17.9}$ & $+16.8$ & $+14.9$ & $\mathbf{+22.1}$ & $+13.2$ & $42{,}237$ & $57$ \\
\midrule
\multirow{2}{*}{Codex} & GPT-6-astra & $\mathbf{54.4}$ & $\mathbf{59.3}$ & $+4.9$ & $+4.5$ & $+5.5$ & $+6.7$ & $-1.4$ & $+9.5$ & $+9.6$ & $2{,}040$ & $14$ \\
 & GPT-5.6-sol & $28.4$ & $35.8$ & $+7.5$ & $+8.0$ & $+6.8$ & $+3.4$ & $+6.5$ & $+6.9$ & $+15.5$ & $6{,}541$ & $18$ \\
\bottomrule
\end{tabular}}
\vspace{-0.1in}
\end{table}

\subsection{Ablation Studies}
\label{sec:ablations}
\begin{figure}[ht]
\centering
\includegraphics[width=\textwidth]{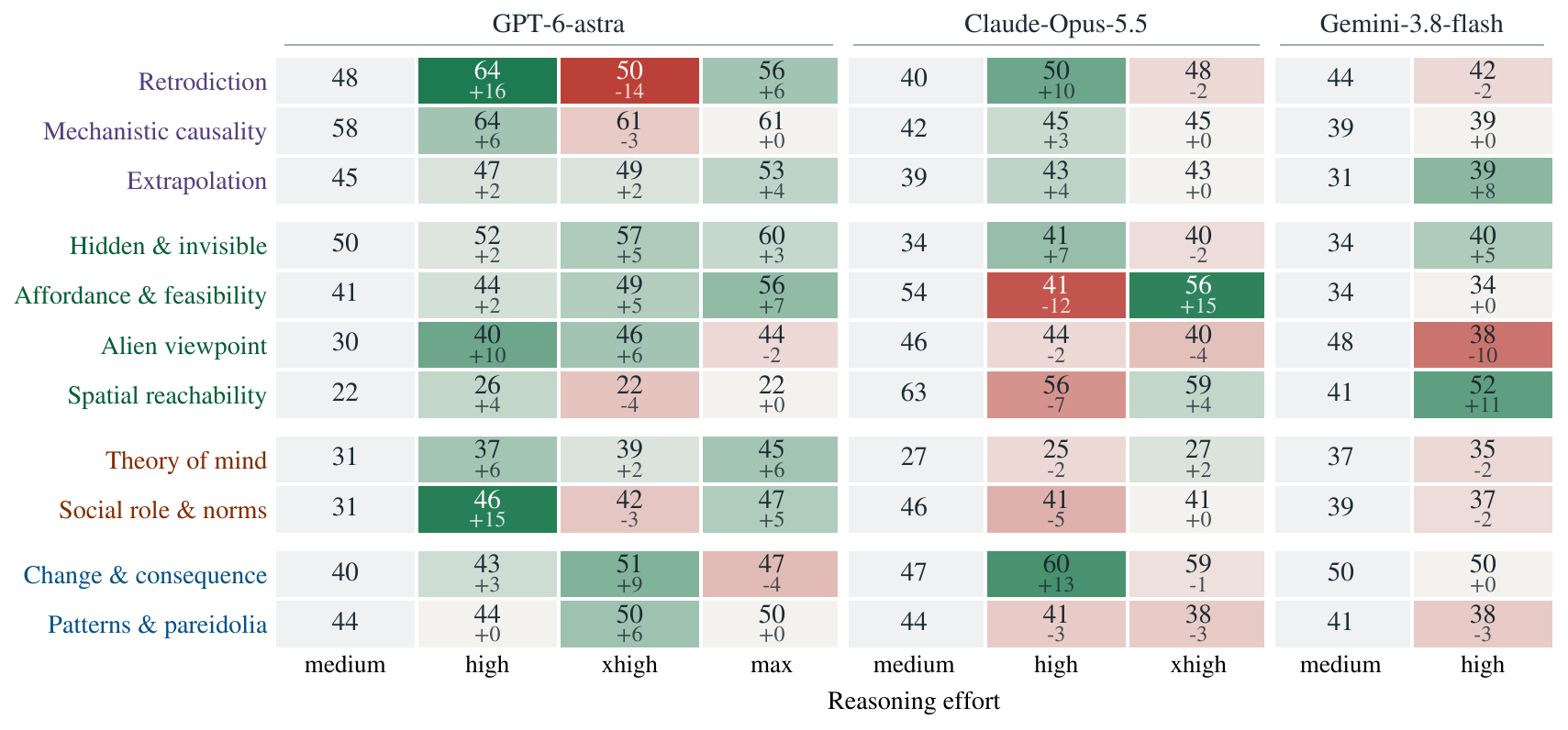}
\caption{\textbf{Performance by subdomains under different reasoning effort} We increase the reasoning effort level from left to right. Green is a gain, red a loss.}
\label{fig:effort_by_domain}
\end{figure}

\paragraph{Effects of Reasoning Effort.}
We evaluate the impact of reasoning effort on model performance across HSS subdomains (Figure~\ref{fig:effort_by_domain}). While increased effort generally improves overall performance, accuracy in certain subdomains degrades at higher reasoning levels. For example, in the retrodiction subdomain, GPT-6-astra exhibits a \textbf{$14$-point drop} when scaling effort from \emph{high} to \emph{xhigh}. Consequently, peak accuracy is frequently achieved at intermediate effort levels rather than the maximum setting. These findings suggest that explicit deliberation is not the primary driver of HSS performance, which aligns with human cognitive observations where similar competencies operate intuitively prior to deliberate reasoning. Extended ablation results are provided in Appendix~\ref{app:effort}.

\paragraph{Effects of Video Frame Budget.} Sweeping the frame budget from $10$ to $500$ across $234$ video tasks shows that performance scales consistently with denser visual sampling. Across four evaluated models, every model achieves its lowest score at $10$ frames, with GPT-6-astra gaining $+17.9$ points when frame budget increases from $10$ to $500$. Extended ablations and per-model results are provided in Appendix~\ref{app:frames}.

\begin{figure}[ht]
\centering
\includegraphics[width=\textwidth]{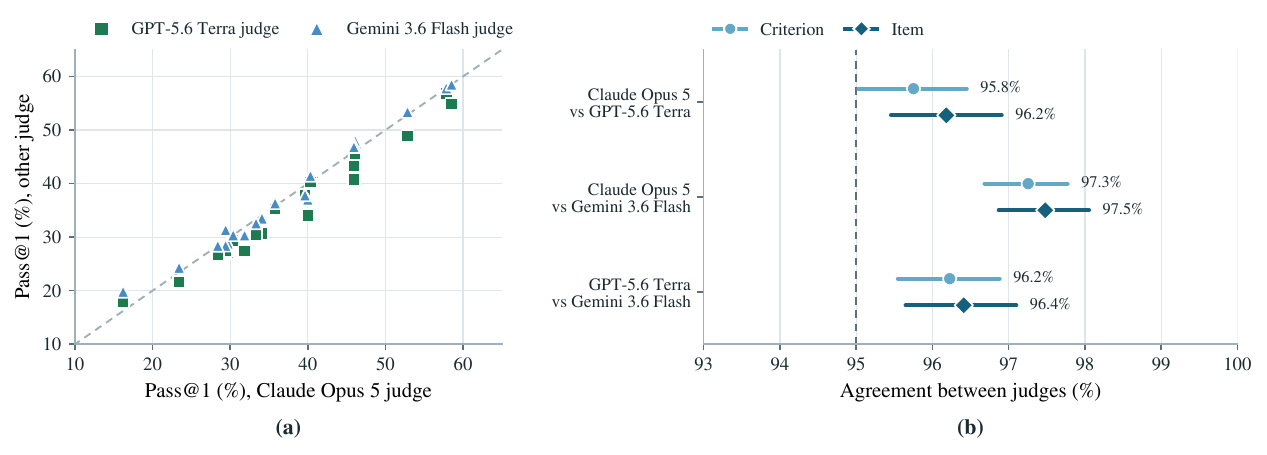}
\caption{\textbf{Judge alignment across vendors.} (a) Pass@1 scores under alternative judges closely track the primary judge (Claude-Opus-5) along the identity line. (b) Pairwise agreement exceeds $95\%$ at both the criterion ($n = 3{,}780$) and the item ($n = 2{,}620$) level for all three judge pairs.}
\label{fig:judge_alignment}
\vspace{-0.1in}
\end{figure}

\paragraph{Effects of different judge models.}
\label{sec:judge_alignment}
To check that the results do not depend on the choice of LLM judge, we re-graded five
models with three judges from three vendors: Claude-Opus-5 (the primary judge),
GPT-5.6-terra and Gemini-3.6-flash, across $2{,}620$ items and $3{,}780$ criteria. Rankings are preserved: judge-to-judge scores correlate at $r \ge 0.98$ (Figure~\ref{fig:judge_alignment}a), and while GPT-5.6-terra is stricter by $2.4$ points on average, it is stricter uniformly. Over $95\%$ of items and
criteria (Figure~\ref{fig:judge_alignment}b), with $\kappa \ge 0.92$ on both units. Finally, there is no self-preference \citep{panickssery2024selfpreference}: Claude-Opus-5 grading itself
shifts its own score by $-0.44$ points. Per-model breakdowns are in Appendix~\ref{app:judge}.

% ===========================================================================
\section{Related Work}
\label{sec:related}
% ===========================================================================

Most current multi-modal benchmarks fall into two families. One wraps problem-solving with visual inputs, such as MMMU, MathVista, Humanity's Last Exam
\citep{yue2024mmmu,lu2024mathvista,phan2025hle,zou2025dynamath}, where the image supplies explicit givens that could in principle be transcribed into text. The other targets perceptual acuity, clutter and content tracking, such as $V^*$, MMBench, Video-MME, BLINK, Perception Test and MVBench \citep{wu2024vstar,liu2024mmbench,fu2024videomme,fu2024blink,patraucean2023perceptiontest,li2024mvbench}, a line that runs back to VQA v2 \citep{goyal2017vqav2}. HSS is orthogonal to both: its answers are neither transcribable nor findable by looking harder, because what the task asks for was never in the frame. 

HSS is also related to a body of visual abduction benchmarks and to spatial, physical,
social and abstract reasoning suites
\citep{zellers2019vcr,hessel2022sherlock,chen2024spatialvlm,riochet2021intphys,xiao2021nextqa,zadeh2019socialiq,sap2019socialiqa,bear2021physion,yi2020clevrer,jin2024mmtomqa,yang2025vsibench,zhang2019raven,kamath2023whatsup}.
Each covers part of this territory, typically one competence at a time and often on
synthetic scenes; HSS spans physical, spatial, social and abstract inference over natural
media under a single admission standard (Appendix~\ref{app:related}). Our agentic condition follows work showing that active inspection, such as cropping, zooming,
re-sampling, sketching can lift visual reasoning
\citep{roberts2025zerobench,guo2025beyond,hu2024sketchpad},
and our grading follows LLM-as-judge practice \citep{zheng2023judging} with an explicit audit of self-preference \citep{panickssery2024selfpreference}
(Section~\ref{sec:judge_alignment}).

\section{Limitations}
\label{sec:limitations}

We note three major limitations. 

\textbf{Visual channel only.} HSS evaluates what a model infers from what it can see. Video items are supplied without their audio track and without a transcript, so items whose resolution would depend on speech or off-screen sound are outside the benchmark's scope by construction.

\textbf{Judge dependence.} Scores are mediated by an LLM judge, which is known to carry biases including self-preference \citep{panickssery2024selfpreference}. We report cross-judge agreement and an oracle condition in which human reference answers are scored correct, which bound the effect.

\textbf{Contamination.} Media sourced from the public web may appear in training data. The questions are newly authored against that media rather than collected with it, which is what the benchmark tests, but prior exposure to an individual image or clip cannot be fully excluded.

% ===========================================================================
\section{Conclusion}
% ===========================================================================
Humans deduce far more from a visual scene than what is explicitly rendered in pixels: a capability measured directly by Humanity’s Sixth Sense (HSS). To benchmark visual intuition objectively, HSS operationalizes it through a structured taxonomy and a rigorous task construction pipeline. Across our benchmarks, frontier models exhibit a stark performance gap compared to the human baseline, a discrepancy that persists despite test-time compute scaling and integration into agentic harnesses. HSS highlights intuitive visual reasoning as a fundamental, measurable axis of multimodal intelligence where current scaling paradigms have so far yielded limited progress.

% ===========================================================================
\subsection*{AI use statement}
% ===========================================================================
AI assistants were used strictly to polish draft and assist in developing evaluation and analysis code. All benchmark tasks, including media selection, questions, reference answers, and grading rubrics were authored and reviewed entirely by human annotators. LLMs appear in this work solely as model baselines of benchmark study and automated evaluators.

\subsection*{Ethics statement}
\textbf{Media licensing.} All media is drawn from three repositories that expose per-item, machine-readable license metadata: Wikimedia Commons, Openverse and the Internet Archive. In addition, only four licenses are admitted: Public Domain, CC0, CC-BY and CC-BY-SA. All four permit redistribution. Any item that is non-commercial, no-derivatives, or whose rights cannot be verified from source metadata is rejected at retrieval time. Creator attribution is carried per item and released with the benchmark, and the source files are mirrored so the frozen benchmark is not exposed to link rot. Furthermore, audio tracks are stripped from all video assets to ensure the benchmark evaluates purely visual reasoning, matching our task design.

\textbf{People in the media.} Many tasks depict identifiable people in public settings.
They are used for research evaluation only, and a keyword screen is applied to video
candidates to avoid material primarily featuring minors. Theory of mind tasks ask models
to infer emotion, intent and belief from appearance and behaviour. Such inferences are
culturally situated and can encode stereotypes. The multi-reviewer consensus requirement
of Section~\ref{sec:design_principles} removes tasks on which readings diverge, but
agreement among a demographically narrow annotator pool is not evidence of a universal
inference. We therefore read theory of mind results as measurements of agreement with
annotator judgement, not as ground truth about the people depicted.

\textbf{Annotation.} Annotators were recruited and compensated through a professional
data-annotation provider at the provider's standard contracted rates.

\subsection*{Reproducibility statement}
We will publicly release our evaluation harness, model registry, and grading prompts. The registry specifies all controlled evaluation parameters, including API endpoints, reasoning effort levels, output token budgets, video processing pathways, and frame caps, which are detailed in Appendix~\ref{app:model_setup} across all $24$ evaluated models and three judges. Where a provider exposes no explicit parameter controls, provider defaults apply. Prompt templates are reproduced verbatim in Appendix~\ref{app:prompts}. All media assets are content-addressed and locally cached to ensure every model receives byte-identical inputs, with each run logging dataset versioning, task media metadata, and per-criterion judge verdicts. We omit sampling temperature overrides to retain vendor defaults across heterogeneous API endpoints. Finally, all confirmatory metrics in this paper are evaluated on a frozen, versioned release containing $522$ tasks (Table~\ref{tab:dataset}).

\bibliography{iclr2027_conference}
\bibliographystyle{iclr2027_conference}

\newpage
\appendix
% ===========================================================================

% ===========================================================================
\section{Extended Related Work}
\label{app:related}
% ===========================================================================

This section expands Section~\ref{sec:related} with the comparisons that place HSS
against individual benchmarks rather than against families of them.

\textbf{Academic problem-solving in a visual wrapper.} MMMU, MathVista, DynaMath and
Humanity's Last Exam \citep{yue2024mmmu,lu2024mathvista,phan2025hle,zou2025dynamath} test high-level
symbolic or domain-specific reasoning over images: a diagram, a plot or a figure
supplies givens, and the difficulty lives in the chain of reasoning applied to them.
The distinguishing property is that a sufficiently complete verbal description of the
image would preserve the task. HSS items do not have this property by measuring the intuitive visual reasoning that does not require deliberate step-by-step derivation.

\textbf{Perception-centred benchmarks.} $V^*$, MMBench, Video-MME, and Video-MME-v2
\citep{wu2024vstar,liu2024mmbench,fu2024videomme,fu2026video} test visual acuity, robustness to clutter, and tracking of content across duration. Here the answer \emph{is} in the
frame, and the question is whether the model can find it. HSS inverts the admission
criterion (Section~\ref{sec:design_principles}): an item is rejected precisely when a
careful enough look would settle it. The two benchmark families are therefore
complementary.

\textbf{Visual abduction and commonsense.} VCR, VisualCOMET and Sherlock
\citep{zellers2019vcr,park2020visualcomet,hessel2022sherlock} are the closest ancestors:
each asks the model to go beyond the depicted scene to intentions, prior events or
likely continuations. They concentrate on social and narrative context and grade
plausible completions, often against multiple-choice alternatives. HSS covers physical,
spatial, social and abstract inference under one taxonomy, and admits only items whose
answer reviewers agree is forced rather than merely plausible, which is what makes
open-ended rubric grading possible.

\textbf{Spatial, physical, and social reasoning.} SpatialVLM, IntPhys, Physion and
CLEVRER \citep{chen2024spatialvlm,riochet2021intphys,bear2021physion,yi2020clevrer}
evaluate relation-naming, physical plausibility and collision prediction, largely over
synthetic or simulated scenes; NExT-QA, Social-IQ, SocialIQa and MMToM-QA
\citep{xiao2021nextqa,zadeh2019socialiq,sap2019socialiqa,jin2024mmtomqa} evaluate causal,
social and mental-state understanding of observed events; VSI-Bench and What's Up
\citep{yang2025vsibench,kamath2023whatsup} probe spatial memory and relational language;
and RAVEN \citep{zhang2019raven} isolates abstract visual analogy. HSS extends both directions to natural media and to inferences
about what is \emph{not} observed: an alien viewpoint the camera never occupies, a rope
whose tension is visible only in its sag, an event that the clip sets up but never
shows.

\textbf{Tool-augmented inspection.} ZeroBench and VisualToolBench
\citep{roberts2025zerobench,guo2025beyond} show that letting a model crop, zoom and
re-sample improves visual reasoning, and motivate our agentic condition. Section~\ref{sec:agentic} uses that lever diagnostically rather than as a method
contribution: if HSS failures were failures of looking, active inspection should repair
them, and the extent to which it does not (Figure~\ref{fig:agentic_failures}) is
evidence about what the benchmark measures.

\textbf{Automated grading and its validity.} Open-ended grading at this scale
requires an LLM judge \citep{zheng2023judging}, which imports known failure modes,
self-preference \citep{panickssery2024selfpreference} foremost among them. We audit
rather than assume: three judges from three vendors, agreement and $\kappa$ per pair,
and a self-grading contrast (Section~\ref{sec:judge_alignment},
Appendix~\ref{app:judge}).

% ===========================================================================
\section{Benchmark Details}
\label{app:HSS}
% ===========================================================================

This appendix documents the released benchmark: what it contains
(Appendix~\ref{app:stats}), how each task was built and reviewed
(Appendix~\ref{app:pipeline}), and how the taxonomy was arrived at
(Appendix~\ref{app:taxonomy}).

\subsection{Composition and statistics}
\label{app:stats}

\begin{figure}[t]
\centering
\includegraphics[width=\textwidth]{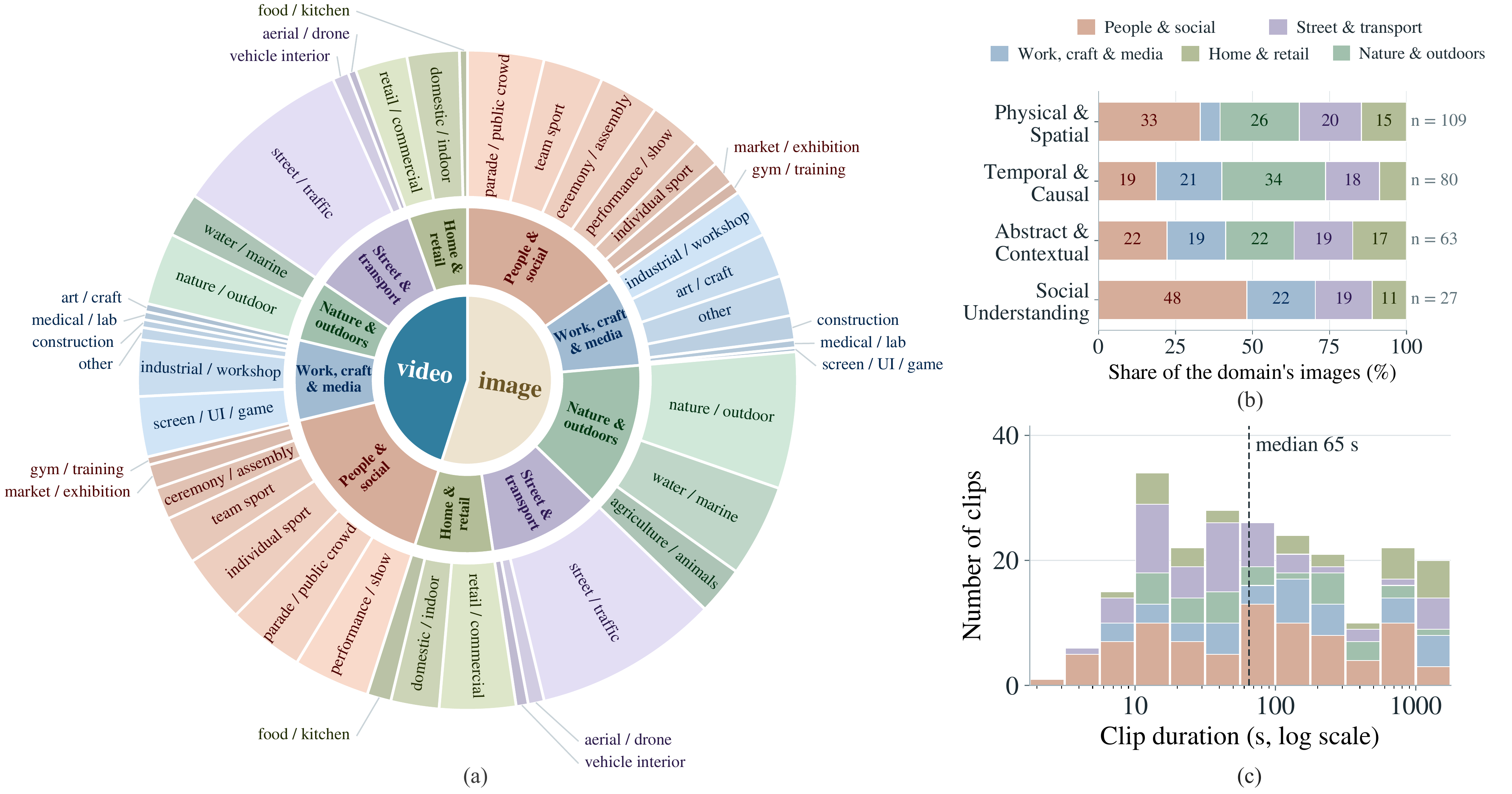}
\caption{\textbf{Media statistics in HSS.} (a)~Hierarchical breakdown of image and video assets across broad visual categories (\emph{People \& social}, \emph{Work, craft \& media}, \emph{Nature \& outdoors}, \emph{Street \& transport}, and \emph{Home \& retail}) and fine-grained sub-scenes. 
(b)~Distribution of visual scene types across the four core reasoning domains, showing diverse environment coverage per domain. 
(c)~Histogram of video clip durations (log scale), spanning a few seconds to $28$ minutes; summary statistics are in Table~\ref{tab:dataset}.}
\label{fig:media_diversity}
\end{figure}

Four properties of the composition in \cref{tab:dataset} matter for the analysis; Figure~\ref{fig:media_diversity} shows the same corpus by scene type and clip length.
First, no domain dominates the benchmark: the largest, \emph{physical \& spatial logic},
holds $176/522$ tasks and the smallest, \emph{abstract \& contextual inference}, holds
$102$, so every domain-level cell is large enough to carry a claim, and the smallest
subdomain still holds $27$. Second, the media split is near-even ($288$ images against
$234$ video clips, $17.6$ hours of video in total), but it is deliberately uneven
\emph{within} domains, because the domains do not need the same evidence: theory of mind
is almost entirely video ($46/51$), since an intention reveals itself over time, whereas
the spatial subdomains lean on images, where a single viewpoint is what makes the
inference hard. Third, the number of criteria per task varies, and $71.5\%$ of tasks
carry exactly one; on those, pass@1 and rubric score coincide by construction, which we
control for in Appendix~\ref{sec:metrics}. Fourth, video length varies by an order of
magnitude across subdomains, so the frame-budget question of Section~\ref{sec:media}
binds very differently in different cells.

\begin{table}[ht]
\caption{\textbf{HSS dataset composition and statistics.} The released benchmark:
$522$ tasks over four domains and eleven subdomains, graded against $723$ rubric
criteria ($1.39$ per task; $71.5\%$ carry exactly one). Each task is paired with a unique
image or video clip, so task counts are also media counts. The video half totals $17.6$
hours: median $76$\,s, interquartile range $20$--$307$\,s, longest $28$\,min. Word counts
are means over whitespace-delimited tokens of the prompt, the full task rubric, and the
reference answer.}
\label{tab:dataset}
\centering
\small
\setlength{\tabcolsep}{4.2pt}
\begin{tabular}{lrrrrrrr}
\toprule
 & \multicolumn{3}{c}{\textbf{Tasks}} & & \multicolumn{3}{c}{\textbf{Mean length (words)}} \\
\cmidrule(lr){2-4}\cmidrule(lr){6-8}
\textbf{Domain / subdomain} & \textbf{All} & \textbf{Image} & \textbf{Video} &
\textbf{Criteria} & \textbf{Prompt} & \textbf{Rubric} & \textbf{Answer} \\
\midrule
{Physical \& spatial logic} & {176} & {114} & {62} & {248} & {37.6} & {26.9} & {73.9} \\
\quad Hidden \& invisible & 58 & 28 & 30 & 83 & 27.7 & 21.8 & 69.7 \\
\quad Spatial ``alien viewpoint'' & 50 & 40 & 10 & 72 & 42.0 & 30.8 & 78.9 \\
\quad Affordance \& feasibility & 41 & 34 & 7 & 60 & 45.4 & 30.0 & 77.8 \\
\quad Spatial reachability & 27 & 12 & 15 & 33 & 39.0 & 25.8 & 68.1 \\
\addlinespace
{Temporal \& causal dynamics} & {134} & {81} & {53} & {173} & {33.2} & {22.2} & {64.4} \\
\quad Extrapolation & 51 & 31 & 20 & 63 & 37.1 & 22.2 & 63.3 \\
\quad Retrodiction & 50 & 40 & 10 & 64 & 33.3 & 21.3 & 73.0 \\
\quad Mechanistic causality & 33 & 10 & 23 & 46 & 27.2 & 23.6 & 53.0 \\
\addlinespace
{Social understanding} & {110} & {28} & {82} & {163} & {29.7} & {28.2} & {64.3} \\
\quad Social role, norm \& power & 59 & 23 & 36 & 93 & 29.1 & 27.7 & 67.1 \\
\quad Theory of mind & 51 & 5 & 46 & 70 & 30.3 & 28.7 & 61.0 \\
\addlinespace
{Abstract \& contextual inference} & {102} & {65} & {37} & {139} & {37.8} & {23.4} & {64.0} \\
\quad Change \& consequence & 70 & 47 & 23 & 84 & 38.1 & 21.9 & 65.4 \\
\quad Patterns \& pareidolia & 32 & 18 & 14 & 55 & 37.2 & 26.8 & 61.1 \\
\midrule
{All of HSS} & {522} & {288} & {234} & {723} & {34.9} & {25.3} & {67.5} \\
\bottomrule
\end{tabular}
\end{table}

\subsection{Construction pipeline}
\label{app:pipeline}

\begin{figure}[ht]
\centering
\includegraphics[width=\textwidth]{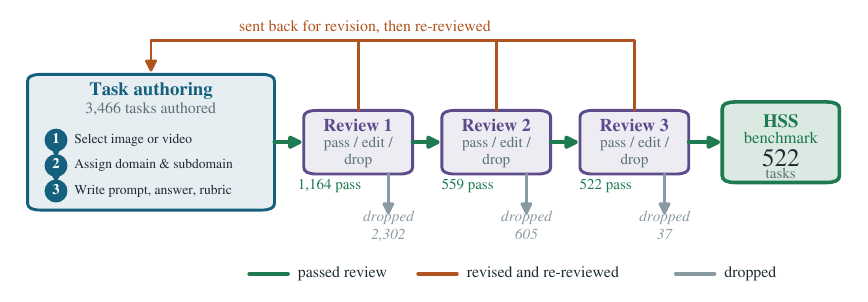}
\caption{\textbf{How an HSS task is built and reviewed}, with the attrition at each
stage. A task is authored in three steps, then faces three independent review rounds;
each round may pass it on, return it to authoring for revision (after which it re-enters
the pipeline from the start), or drop it outright. Counts are net survivors: a task
returned for revision and later re-admitted is counted once, at the round it finally
passes, so each \emph{dropped} figure is the net loss at that round. Of $3{,}466$
authored tasks, $522$ survive all three rounds, a yield of $15.1\%$. The first gate is
by far the steepest: it turns away $2{,}302$ tasks, two in three of everything written,
overwhelmingly because the answer was already depicted in the scene, or for want of
unambiguous consensus on the ground truth (principles 1 and 3 of
Section~\ref{sec:design_principles}).}
\label{fig:pipeline}
\end{figure}

Figure~\ref{fig:pipeline} diagrams the authoring and review pipeline of
Section~\ref{sec:construction} end to end, with the number of tasks surviving each
round. We include it because the attrition is itself a description of the benchmark:
the three principles of Section~\ref{sec:design_principles} are cheap to state and
expensive to enforce, and the cost shows up as the $85\%$ of authored tasks that do not
reach the release.

\subsection{Taxonomy design}
\label{app:taxonomy}

Table~\ref{tab:hss-taxonomy} gives the taxonomy in full: each domain with the kind of
recovery it demands, and each subdomain with the inferential step that defines it. This
section says where the categories came from and how an item is assigned to one.

\paragraph{Four main domains partition.} Temporal \& causal dynamics withholds another
moment in time: the scene is one slice, and the answer lies behind it, ahead of it, or
in the mechanism carrying one to the other. Physical \& spatial logic withholds
structure: what sits outside the frame, what an object weighs or how it would move, how
the layout resolves from a viewpoint the camera never occupied. Social understanding
withholds mental and normative state: intent, belief, role, the rule everyone in the
scene is following. Abstract \& contextual inference withholds the rule itself: the
convention, pattern or consequence that makes the arrangement mean something. The split
follows the lines along which cognitive science studies these intuitive inferences separately: core knowledge of objects and space \citep{spelke2007core}, intuitive physics
\citep{battaglia2013simulation}, and mentalizing \citep{baker2017rational}.

\begin{table}[ht]
\caption{Taxonomy Design for HSS along with descriptions.}
\label{tab:hss-taxonomy}
\centering
\scriptsize
\begin{tabularx}{\textwidth}{@{} p{2.2cm} >{\hsize=0.95\hsize}X p{2.2cm} >{\hsize=1.05\hsize}X @{}}
\toprule
\textbf{Domain} & \textbf{Domain Description} & \textbf{Sub-domain} & \textbf{Sub-domain Description} \\
\midrule

\multirow{7}{=}{Temporal \& Causal Dynamics} 
  & \multirow{7}{=}{Recovering the scene's state at a moment other than the one depicted, or the causal process connecting those moments. The media is one slice; the answer lies behind it, ahead of it, or in the mechanism driving it.}
  & Retrodiction & Inferring what already happened based on visual traces. \\
  \cmidrule(l){3-4}
  & & Mechanistic Causality & Identifying the immediate physical mechanism driving an ongoing event. This is about understanding the ``how'' and ``why'' of current motion or state changes, not predicting the future. \\
  \cmidrule(l){3-4}
  & & Extrapolation & Predicting the immediate kinematic trajectory or long-term systemic outcome of the current scene. \\
\midrule

\multirow{10}{=}{Physical \& Spatial Logic} 
  & \multirow{10}{=}{Recovering unshown structure of the physical scene: what exists outside view, what properties objects hold beyond their appearance, and how the layout resolves from a coordinate the camera never occupied.}
  & Hidden \& Invisible & Reasoning about entities or physical properties that are not directly visible in a scene. Deducing occluded, off-screen, or internal objects, non-visual attributes (mass, friction), and invisible forces. \\
  \cmidrule(l){3-4}
  & & Affordance \& Feasibility & Assessing the physical compatibility between objects or between agents and their environment. \\
  \cmidrule(l){3-4}
  & & Spatial ``Alien Viewpoint'' & Mentally re-projecting the 3D scene to calculate line-of-sight or reachability from a different physical coordinate. \\
  \cmidrule(l){3-4}
  & & Spatial Reachability & Evaluating accessible space, paths, and physical volumes an agent or object can extend to within the 3D layout. \\
\midrule

\multirow{4}{=}{Social Understanding} 
  & \multirow{4}{=}{Recovering the unobservable mental and normative layer: what agents believe, want, feel, attend to, and fail to know --- and what the situation's unwritten rules permit, forbid, or make awkward.}
  & Theory of Mind & Reasoning about the internal beliefs, knowledge gaps, deceptions, and emotional states of an agent. \\
  \cmidrule(l){3-4}
  & & Social Role, Norm \& Power Dynamics & Reasoning about unwritten social contracts, etiquette, and cultural appropriateness in a scene. \\
\midrule

\multirow{5}{=}{Abstract \& Contextual Inference} 
  & \multirow{5}{=}{Recovering what a scene implies but never shows: the conventions and circumstances that give it meaning, and the outcomes of worlds it doesn't depict. The answer follows from applying a rule or reading a context.}
  & Change \& Consequence & Reasoning about a change the scene doesn't contain: given the change, inferring its consequence; or given the consequence, inferring the change required. \\
  \cmidrule(l){3-4}
  & & Patterns \& Pareidolia & Identifying structure in random data or showing specific hardwired bias to detect familiar patterns---especially faces and human figures---in random noise or objects. \\
\bottomrule
\end{tabularx}
\end{table}

% ===========================================================================
\section{Experimental Setup}
\label{app:exp_setup}
% ===========================================================================

This section details the experimental setup.

\subsection{Model configurations}
\label{app:model_setup}

Table~\ref{tab:model_setup} shows how every model in
Table~\ref{tab:main} was called. It is generated from the harness's model registry (\texttt{configs/models.yaml}), which is released with the evaluation code, so a reader can reproduce any row from the endpoint and the settings given here.

Several conventions apply to every row and are therefore not repeated in the table. \emph{Temperature} and \emph{top-p} are never set: we omit both from the request so that each provider's own default applies. \emph{Reasoning effort} is run at the highest level each deployment honours when possible. The effort listed is the one used for the main table. \emph{Max output tokens} is the requested ceiling, not the observed usage; thinking tokens count against it.

For video, models advertising native ingestion receive the clip as a single content block. Everything else receives uniformly sampled frames at $2$~fps, each labelled with its timestamp, capped at the per-model frame limit in the last column: $500$ unless a provider enforces something stricter, in which case frames are thinned uniformly with the endpoints
retained, so the frames still span the whole clip. Frames are downscaled to a longest side of $768$~px and encoded at a fixed JPEG quality; images are sent at native resolution with a longest side of $4096$~px and are downscaled only when a payload limit requires it. Requests carry a $1200$~s timeout with up to $5$ retries under exponential backoff.

\begin{table}[t]
\caption{\textbf{Model configurations for the main experiments.} All $25$ models of Table~\ref{tab:main}, grouped by vendor in the same order, plus the three judges of Section~\ref{sec:judge_alignment}. \emph{Endpoint} is the identifier sent to the proxy; \emph{API} is the request shape (OpenAI Chat Completions, OpenAI Responses, or Anthropic Messages). \emph{Effort} is the reasoning level used for the main table. \emph{Out} is the requested output-token ceiling, which thinking tokens count against. \emph{Native vid.} is whether the clip is sent as a video block rather than as sampled frames. \emph{Frames} is the per-request frame cap that binds for that deployment; sampling is $2$~fps for all models, thinned uniformly to fit the cap.}
\label{tab:model_setup}
\centering
\small
\setlength{\tabcolsep}{4pt}
\resizebox{\linewidth}{!}{
\begin{tabular}{lllllrcr}
\toprule
\textbf{Vendor} & \textbf{Model} & \textbf{Endpoint} & \textbf{API} & \textbf{Effort} &
\textbf{Out} & \textbf{Native vid.} & \textbf{Frames} \\
\midrule
\multirow{7}{*}{OpenAI} & GPT-6-astra & \texttt{openai/gpt-6-astra} & Responses & \texttt{max} & $128,000$ & no & $500$ \\
 & GPT-6.1-sol & \texttt{openai/gpt-6.1-sol} & Responses & \texttt{max} & $128,000$ & no & $500$ \\
 & GPT-6-sol & \texttt{openai/gpt-6-sol} & Responses & \texttt{max} & $128,000$ & no & $500$ \\
 & GPT-5.6-sol & \texttt{openai/gpt-5.6-sol} & Responses & \texttt{max} & $128,000$ & no & $500$ \\
 & GPT-5.6-terra & \texttt{openai/gpt-5.6-terra} & Responses & \texttt{max} & $128,000$ & no & $500$ \\
 & GPT-5.6-luna & \texttt{openai/gpt-5.6-luna} & Responses & \texttt{max} & $128,000$ & no & $500$ \\
 & GPT-6-luna & \texttt{openai/gpt-6-luna} & Responses & \texttt{max} & $128,000$ & no & $500$ \\
\addlinespace
\multirow{6}{*}{Anthropic} & Claude-Opus-5.5 & \texttt{anthropic/claude-opus-5-5} & Messages & \texttt{xhigh} & $128,000$ & no & $500$ \\
 & Claude-Fable-5.1 & \texttt{anthropic/claude-fable-5-1} & Messages & \texttt{xhigh} & $128,000$ & no & $500$ \\
 & Claude-Fable-5 & \texttt{anthropic/claude-fable-5} & Messages & \texttt{max} & $128,000$ & no & $500$ \\
 & Claude-Opus-5 & \texttt{anthropic/claude-opus-5} & Messages & \texttt{max} & $128,000$ & no & $500$ \\
 & Claude-Sonnet-5 & \texttt{anthropic/claude-sonnet-5} & Messages & \texttt{max} & $128,000$ & no & $500$ \\
 & Claude-Opus-4.8 & \texttt{anthropic/claude-opus-4-8} & Messages & \texttt{max} & $128,000$ & no & $500$ \\
\addlinespace
\multirow{4}{*}{Google} & Gemini-3.8-flash & \texttt{gemini/gemini-3.8-flash} & Chat & \texttt{high} & $65,536$ & yes & $500$ \\
 & Gemini-3.7-flash & \texttt{gemini/gemini-3.7-flash} & Chat & \texttt{high} & $65,536$ & yes & $500$ \\
 & Gemini-3.6-flash & \texttt{gemini/gemini-3.6-flash} & Chat & \texttt{high} & $16,384$ & yes & $500$ \\
 & Gemini-3.5-flash & \texttt{gemini/gemini-3.5-flash} & Chat & \texttt{high} & $16,384$ & yes & $500$ \\
\addlinespace
\multirow{2}{*}{Meta} & Muse Spark 1.3 & \texttt{meta\_ai/muse-spark-1.3} & Responses & \texttt{max} & $128,000$ & yes & $50$ \\
 & Muse Spark 1.2 & \texttt{meta\_ai/muse-spark-1.2} & Responses & \texttt{xhigh} & $128,000$ & yes & $50$ \\
\addlinespace
\multirow{2}{*}{Alibaba} & Qwen3.8-Max & \texttt{dashscope/qwen3.8-max} & Chat & \texttt{max} & $16,384$ & no & $250$ \\
 & Qwen3.8-flash & \texttt{dashscope/qwen3.8-flash} & Chat & \texttt{max} & --- & --- & --- \\
\addlinespace
\multirow{1}{*}{Zhipu} & GLM-5.3-flash & \texttt{fireworks\_ai/glm-5p3-flash} & Chat & \texttt{max} & $128,000$ & no & $60$ \\
\addlinespace
\multirow{1}{*}{Moonshot} & Kimi-K3 & \texttt{bedrock/us.moonshotai.kimi-k3} & Chat & \texttt{max} & $128,000$ & no & $100$ \\
\addlinespace
\multirow{2}{*}{Xiaomi} & MiMo-v2.6-flash & \texttt{mimo-v2.6-flash} & Chat & \texttt{high} & --- & --- & --- \\
 & MiMo-v2.6-pro & \texttt{mimo-v2.6-pro} & Chat & \texttt{high} & --- & --- & --- \\
\midrule
\multirow{3}{*}{Judges} & judge-claude-opus-5 & \texttt{anthropic/claude-opus-5} & Chat & \texttt{high} & $8,192$ & n/a & n/a \\
 & judge-gpt-5.6-terra & \texttt{openai/gpt-5.6-terra} & Chat & \texttt{high} & $8,192$ & n/a & n/a \\
 & judge-gemini-3.6-flash & \texttt{gemini/gemini-3.6-flash} & Chat & \texttt{high} & $8,192$ & n/a & n/a \\
\bottomrule
\end{tabular}}
\end{table}

\subsection{Media handling}
\label{sec:media}

\paragraph{Native video.} Models that accept a video file receive the clip directly. 

\paragraph{Uniform frame sampling.} Otherwise, videos are sampled at a fixed rate of $f = 2$~fps with a hard cap of $N_{\max}$ frames. When $f \cdot T > N_{\max}$ for a clip of duration $T$, we \emph{lower the rate} to $N_{\max}/T$ rather than truncating, so the frames always span the whole clip:
\begin{equation}
f_{\text{eff}} = \min\!\left(f,\ \frac{N_{\max}}{T}\right), \qquad
N = \min\!\left(\lfloor f T \rfloor,\ N_{\max}\right).
\end{equation}
Each frame is labelled with its timestamp, and the prompt states the frame count, effective rate and duration, so the model can reason about elapsed time rather than mere ordering. We use $N_{\max} = 500$ as a single universal cap for all models, chosen as the most restrictive per-request image limit among the deployments we evaluate; where a provider imposes a stricter limit, frames are thinned \emph{uniformly} (endpoints retained) and the affected samples are flagged so the reduced temporal resolution is visible in the analysis rather than hidden.

\paragraph{Images.} Sent at native resolution, downscaled only when a payload limit requires it. All media is content-addressed and cached, so every model sees byte-identical input.

\subsection{Judging}

A single LLM judge grades all runs, following standard practice for free-form evaluation \citep{zheng2023judging}. The judge sees the question, the reference answer, the rubric, and the candidate answer, and returns a per-criterion verdict with a one-sentence justification. It does \emph{not} see the media by default: rubrics are written to be checkable against the reference answer, which keeps the judge from re-adjudicating perception. We report the effect of changing judge model in Section~\ref{sec:judge_alignment}.

\paragraph{Grading protocol.} The judge is instructed to: mark a criterion met only if the answer actually asserts what it requires; mark a criterion unmet if the answer does not state the required content; ignore extra correct detail; and treat the reference answer as ground truth on disagreement. It returns one verdict per criterion with a short justification. Unparseable output is re-requested a bounded number of times before the sample is recorded as a grading failure rather than a zero. The system prompt and output schema are in Appendix~\ref{app:prompts}.

\subsection{Metrics}
\label{sec:metrics}

Every number reported in this paper is \textbf{pass@1}. Specifically, each sample $i \in \{1, \dots, n\}$ carries $k_i$ atomic rubric criteria and is attempted $A_i$ times: $A_i = 3$ for every model in Table~\ref{tab:main}, except where a provider error left fewer usable attempts, and $A_i = 1$ in the reasoning-effort and frame-budget sweeps (Appendices~\ref{app:effort} and~\ref{app:frames}), where pass@1 reduces to the observed pass rate. Let $v_{ija} \in \{0,1\}$ be the judge's verdict on criterion $j$ of sample $i$
at attempt $a$. An \emph{attempt} counts as correct only when every criterion is met,
\begin{equation}
c_{ia} = \mathbbm{1}\!\left[\textstyle\sum_{j=1}^{k_i} v_{ija} = k_i\right],
\end{equation}
so a response that supplies two of three required facts is not a correct answer to the question asked. Pass@1 is then the fraction of attempts that are correct, averaged over
samples:
\begin{equation}
\text{pass@1} = \frac{1}{n}\sum_{i=1}^{n} \underbrace{\frac{1}{A_i}\sum_{a=1}^{A_i} c_{ia}}_{\text{per-sample pass@1}},
\qquad
\text{Rubric score} = \frac{1}{n}\sum_{i=1}^{n}\frac{1}{A_i}\sum_{a=1}^{A_i}\frac{1}{k_i}\sum_{j=1}^{k_i} v_{ija}.
\end{equation}
Read as an estimator, the per-sample term is the probability that one sampled response solves that task, and the outer mean weights every sample equally regardless of how many
attempts it has. This is the $k = 1$ case of the pass@$k$ estimator used for code generation \citep{chen2021codex}; because $k = 1$, no combinatorial correction is needed
and the estimate is simply the observed success rate.

\subsection{The blind control}
\label{app:blind_test_example}

The blind control re-runs the strongest model, GPT-6-astra at maximum effort, on all $522$ tasks with the media removed and everything else held fixed (such as prompt, rubrics, and judge). It scores $6.6\%$ (Table~\ref{tab:main}) against $53.6\%$ sighted. Only $4$ of the $522$ tasks pass blind on all three attempts, while $67$ pass at least once: the blind model is not solving a stable subset, it is landing on two-way questions at roughly the rate chance predicts. Below we show four examples pass on the first attempt and fail on at least one of the other two.

\begin{promptbox}[promptgrey]{What passes with the media removed}
\small
\textbf{1. A two-way question.} ``Assuming the person seated in the rowboat was rowing in
a constant direction, did they pass in front of or behind the boat on the left?''
Blind answer, in full: ``They passed behind the boat on the left.'' Correct, with no
stated reason, and it flips on attempts 2 and 3.\\[3pt]
\textbf{2. A yes/no question.} ``Without standing up, could the musician closest to the
left of the director reach out and touch the director with his hand?'' Blind answer:
``No. The director is too far away for the seated musician to touch him without getting
up.'' The reference says no, so the rubric's single criterion is met by a coin
flip dressed in a justification the blind model cannot have.\\[3pt]
\textbf{3. A left/right question.} ``\ldots from which of his sides did the puck pass
through?'' Blind answer: ``The puck passed on his left side.'' The sighted model gets
this one wrong on all three attempts, which is a fair summary of how little the blind
pass means.\\[3pt]
\textbf{4. A question that carries its own answer.} ``What recent physical process most
likely created the muddy channel that the rescue workers are clearing, and what visible
evidence supports that conclusion?'' Blind answer: ``scoured by a recent flash flood or
mud-and-debris flow\ldots torn-away banks with exposed roots, uprooted trees, broken
logs''. The words \emph{muddy channel} and \emph{rescue workers} narrow the process
before the image is consulted, and the model then invents evidence that happens to match.
\end{promptbox}

\subsection{Prompts}
\label{app:prompts}

Below we show the prompts used for HSS evaluations.

\begin{promptbox}{Media preamble --- image}
\small\ttfamily
An image is attached.
\end{promptbox}

\begin{promptbox}{Media preamble --- native video}
\small\ttfamily
A video (\{duration\}s) is attached. Reason about motion and event order across time.
\end{promptbox}

\begin{promptbox}{Media preamble --- sampled frames}
\small\ttfamily
The video below is provided as \{num\_frames\} frames sampled uniformly at \{fps\} frames per second (video duration \{duration\}s), in chronological order. Frame timestamps in seconds: \{timestamps\}.\\ Treat the frames as one continuous video and reason about motion and event order across them.
\end{promptbox}

\begin{promptbox}[promptgrey]{Blind control system prompt (no media attached)}
\small\ttfamily
You are a helpful assistant. Answer the user's question faithfully. If you do not have enough information to answer, do not guess --- say so honestly.
\end{promptbox}

\paragraph{Judge prompt.} The judge receives the question, the reference answer, the numbered rubric and the candidate answer, and returns one verdict per criterion in a fixed JSON schema. It does \emph{not} see the media: rubrics are written to be checkable against the reference answer, which keeps the judge from re-adjudicating perception.

\begin{promptbox}[promptgreen]{Judge system prompt}
\small\ttfamily
You are a strict, impartial grader for the Humanity's Sixth Sense (HSS) visual reasoning
benchmark. You are given a question about a piece of media, a human-written reference
answer that is authoritative and correct, a list of rubric criteria, and a candidate
answer from a model under evaluation.\\[3pt]
Grade the candidate answer against each rubric criterion independently.\\[3pt]
Rules:\\
1. A criterion is met only if the candidate answer actually asserts what the criterion
requires. Do not infer or give credit for near-misses, hedges, or contradictions.\\
2. Wording may differ from the reference; judge meaning, not phrasing. Synonyms,
equivalent directions (e.g.\ 'right to left' vs 'leftward'), and equivalent units count.\\
3. If the candidate answer states the required content but also states a contradicting
alternative, the criterion is NOT met.\\
4. Extra correct detail neither helps nor hurts. Extra incorrect detail only matters if
it contradicts a criterion.\\
5. Treat the reference answer as ground truth when the candidate disagrees with it.\\
6. Judge only the criteria given. Do not invent criteria.\\[3pt]
Return ONLY a JSON object, no prose and no markdown fences, in exactly this form:\\
\{"criteria": 
[\{"id": "<criterion id>", "met": true|false, "justification": "<one short
sentence citing the candidate answer>"\}], "overall\_comment": "<one sentence overall
assessment>"\}
\end{promptbox}

% ===========================================================================
\section{Failure Analysis}
\label{app:failures}
% ===========================================================================

\paragraph{Taxonomy and protocol.} We label the primary cause of every failed first attempt in the main table: $8{,}573$ failures over $522$ tasks and the $24$ models evaluated when the analysis was run, which is every row of Table~\ref{tab:main} except the newly added GPT-6.1-sol. The
eleven causes fall into three families that separate \emph{where} the answer goes wrong: 
\begin{itemize}
    \item \textbf{Perception}: where the model misreads something the scene actually shows for implicit visual reasoning.
    \item \textbf{Latent}: where it misses or misreads something the scene only implies.
    \item \textbf{Reasoning}: where it has the facts right and draws the wrong conclusion.
\end{itemize}

A twelfth label, \emph{no committed answer}, covers non-answers. Labelling is performed by Claude-Opus-5 shown the question, the reference answer and the model's answer. Every failure received a label. Table~\ref{tab:failure_causes} gives the pooled distribution on failure annotation.

\begin{table}[h]
\caption{\textbf{Primary failure causes, pooled over $24$ models} ($n = 8{,}573$).
\emph{Share} is the percentage of all failures; the last two columns split that share by
media type. Perception and latent causes together account for $94\%$ of failures.}
\label{tab:failure_causes}
\centering
\small
\begin{tabular}{llrrrr}
\toprule
\textbf{Family} & \textbf{Cause} & \textbf{Count} & \textbf{Share} &
\textbf{Image} & \textbf{Video} \\
\midrule
Perception & Misidentified an object, person, or role & $1,702$ & $19.9$ & $17.7$ & $22.2$ \\
 & Misread orientation or facing & $1,031$ & $12.0$ & $18.2$ & $5.1$ \\
 & Misjudged size, distance, or fit & $1,048$ & $12.2$ & $15.8$ & $8.2$ \\
 & Read a 2D overlap as 3D alignment & $776$ & $9.1$ & $14.6$ & $2.9$ \\
Latent & Misread an invisible quantity from its trace & $634$ & $7.4$ & $6.9$ & $8.0$ \\
 & Misread motion or sequence & $678$ & $7.9$ & $3.8$ & $12.4$ \\
 & Misread intent, feeling, or relationship & $373$ & $4.4$ & $0.3$ & $8.9$ \\
 & Missed the decisive cue & $1,826$ & $21.3$ & $15.2$ & $28.1$ \\
Reasoning & Converted to the wrong viewpoint & $120$ & $1.4$ & $1.7$ & $1.1$ \\
 & Misjudged a line of sight & $219$ & $2.6$ & $3.7$ & $1.3$ \\
 & Mis-traced the consequence or the question's condition & $115$ & $1.3$ & $2.1$ & $0.5$ \\
--- & No committed answer & $51$ & $0.6$ & $0.1$ & $1.2$ \\
\bottomrule
\end{tabular}
\end{table}

\paragraph{The cause follows the task.} Cause is strongly associated with subdomain and with media type, but almost not at all with which model produced the failure. The per-task view agrees: on the $503$ tasks that five or more models fail, a median $80\%$ of those models fail for the same cause and $90\%$ for the same family. Figure~\ref{fig:failure_causes} gives the full subdomain $\times$ cause matrix.

The media split is the clearest single contrast. Image failures are $66\%$ perception, dominated by misread orientation ($18\%$), misidentification ($18\%$), size and fit
($16\%$) and reading a 2D overlap as 3D alignment ($15\%$). Video failures invert this: $57\%$ are latent, with the missed decisive cue alone at $28\%$. Video does not make
models worse at seeing; it gives them more to miss.

\paragraph{What the agent loop repairs.} Table~\ref{tab:failure_shift} pairs each agentic run against its own single-call run on the same tasks, pooling the four
harness/base pairs of Table~\ref{tab:agentic_full}. Every perception and latent cause falls by $19$--$37\%$, with one
exception: reading a 2D overlap as 3D alignment does not move ($102 \rightarrow 101$), and only $24\%$ of single-call failures of that kind are repaired, the lowest fix
rate of any common perception cause. Cropping and zooming recovers a missed cue or a misidentified object; it does not recover depth, because the extra looking returns more of the same projection.

\begin{table}[h]
\caption{\textbf{How the agent loop shifts the failure distribution}, four
harness/base pairs pooled on $387$ paired tasks, first attempts. \emph{Fixed} is the
percentage of single-call failures with that cause that the harness repairs.}
\label{tab:failure_shift}
\centering
\small
\begin{tabular}{llrrrr}
\toprule
\textbf{Family} & \textbf{Cause} & \textbf{Single} & \textbf{Agentic} &
\textbf{$\Delta$ \%} & \textbf{Fixed \%} \\
\midrule
Perception & Misidentified an object, person, or role & $181$ & $147$ & $-19$ & $35$ \\
Perception & Misread orientation or facing & $117$ & $93$ & $-21$ & $40$ \\
Perception & Misjudged size, distance, or fit & $137$ & $105$ & $-23$ & $39$ \\
Perception & Read a 2D overlap as 3D alignment & $102$ & $101$ & $-1$ & $24$ \\
Latent & Misread an invisible quantity from its trace & $71$ & $55$ & $-23$ & $37$ \\
Latent & Misread motion or sequence & $83$ & $52$ & $-37$ & $33$ \\
Latent & Misread intent, feeling, or relationship & $49$ & $35$ & $-29$ & $31$ \\
Latent & Missed the decisive cue & $154$ & $110$ & $-29$ & $34$ \\
Reasoning & Converted to the wrong viewpoint & $13$ & $14$ & $+8$ & $31$ \\
Reasoning & Misjudged a line of sight & $22$ & $27$ & $+23$ & $9$ \\
Reasoning & Mis-traced the consequence or the question's condition & $12$ & $15$ & $+25$ & $25$ \\
--- & No committed answer & $3$ & $5$ & $+67$ & $33$ \\
\bottomrule
\end{tabular}
\end{table}

\begin{table}[h]
\caption{\textbf{Where the agent loop pays, by subdomain.} Pass@1 (\%) for the four
harness/base pairs pooled on the $387$ paired tasks of Table~\ref{tab:failure_shift},
first attempts. $n$ counts pair $\times$ task observations, so each subdomain appears
four times. Re-inspection helps most where the evidence is in the frame and least where
the answer turns on depth or on an outcome the media never shows.}
\label{tab:agentic_subdomain}
\centering
\small
\begin{tabular}{lrrrr}
\toprule
\textbf{Subdomain} & \textbf{$n$} & \textbf{Single} & \textbf{Agentic} & \textbf{$\Delta$} \\
\midrule
Patterns \& pareidolia & $68$ & $33.8$ & $64.7$ & $\mathbf{+30.9}$ \\
Retrodiction & $152$ & $34.9$ & $55.3$ & $+20.4$ \\
Theory of mind & $136$ & $32.4$ & $48.5$ & $+16.2$ \\
Social role, norm \& power & $168$ & $32.1$ & $47.6$ & $+15.5$ \\
Affordance \& feasibility & $140$ & $43.6$ & $55.0$ & $+11.4$ \\
Change \& consequence & $224$ & $40.6$ & $49.1$ & $+8.5$ \\
Hidden \& invisible & $180$ & $40.6$ & $48.9$ & $+8.3$ \\
Mechanistic causality & $84$ & $47.6$ & $56.0$ & $+8.3$ \\
Spatial ``alien viewpoint'' & $152$ & $42.1$ & $50.0$ & $+7.9$ \\
Extrapolation & $160$ & $45.0$ & $52.5$ & $+7.5$ \\
Spatial reachability & $84$ & $34.5$ & $39.3$ & $+4.8$ \\
\bottomrule
\end{tabular}
\end{table}

% ===========================================================================
\section{Failure Examples}
\label{app:subdomain_cases}
% ===========================================================================

\begin{figure}[p]
\centering
\includegraphics[width=\textwidth]{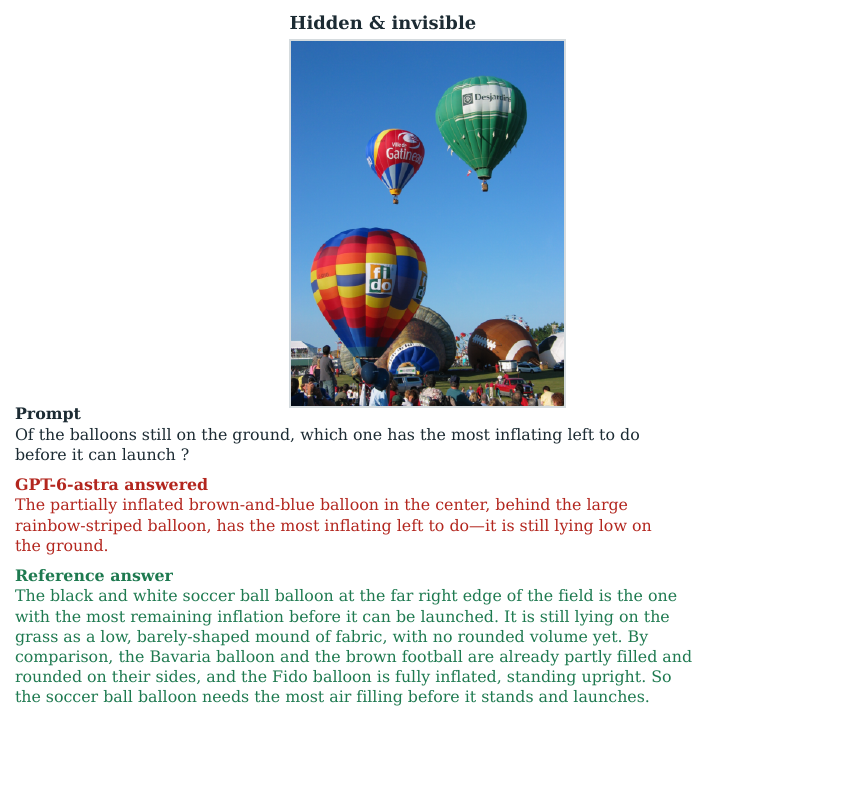}
\caption{\textbf{Failure example: hidden \& invisible}. The model's answer and the human-written reference answer, with the media the
model was served.}
\label{fig:failure_example_2}
\end{figure}

\begin{figure}[p]
\centering
\includegraphics[width=\textwidth]{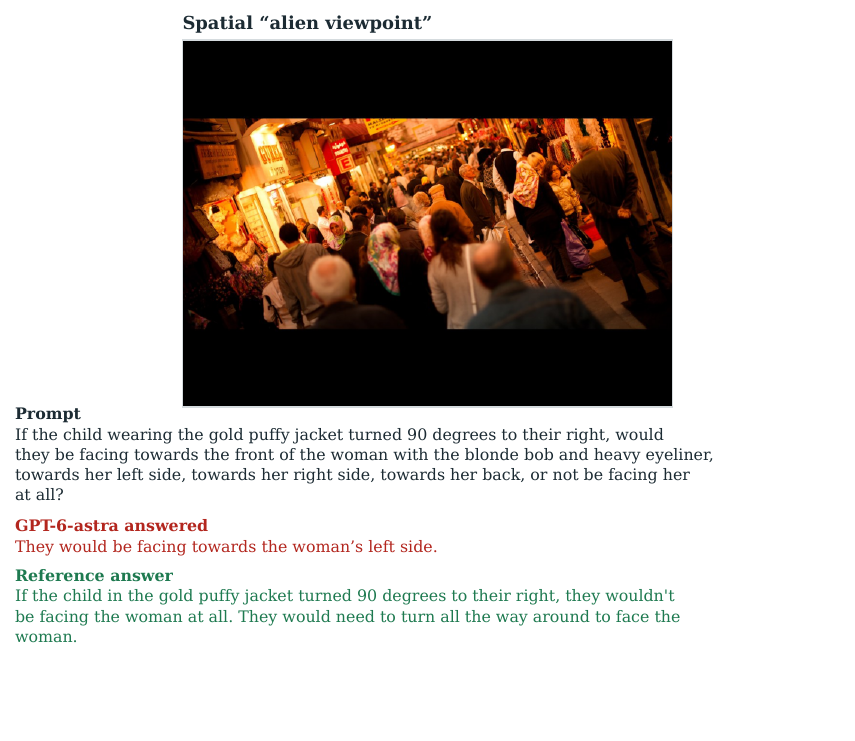}
\caption{\textbf{Failure example: the spatial ``alien viewpoint''}. The model's answer and the human-written reference answer, with the media the
model was served.}
\label{fig:failure_example_3}
\end{figure}

\begin{figure}[p]
\centering
\includegraphics[width=\textwidth]{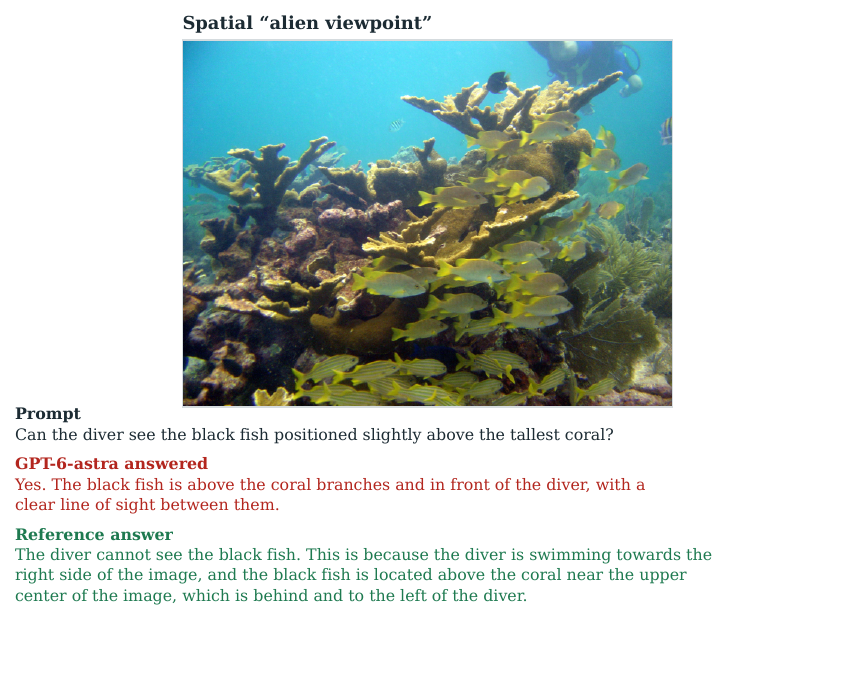}
\caption{\textbf{Failure example: the spatial ``alien viewpoint''}. The model's answer and the human-written reference answer, with the media the
model was served.}
\label{fig:failure_example_4}
\end{figure}

\begin{figure}[p]
\centering
\includegraphics[width=\textwidth]{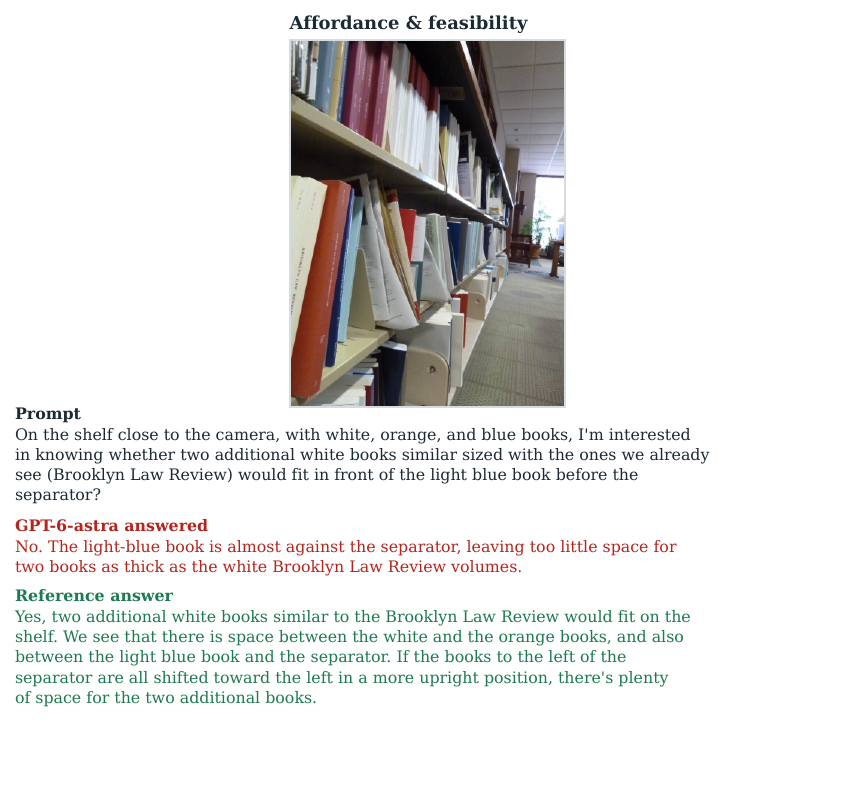}
\caption{\textbf{Failure example: affordance \& feasibility}. The model's answer and the human-written reference answer, with the media the
model was served.}
\label{fig:failure_example_5}
\end{figure}

\begin{figure}[p]
\centering
\includegraphics[width=\textwidth]{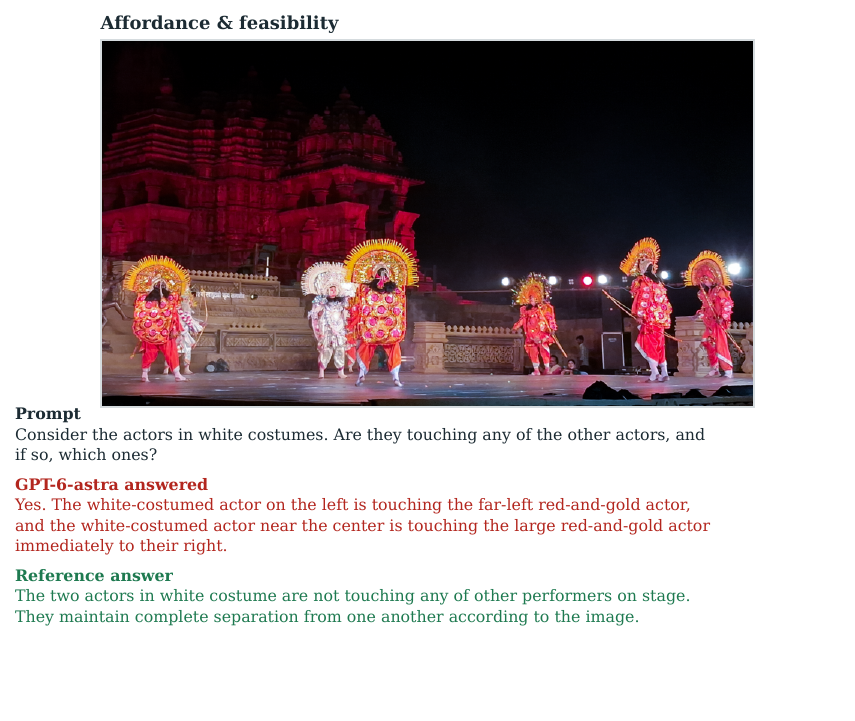}
\caption{\textbf{Failure example: affordance \& feasibility}. The model's answer and the human-written reference answer, with the media the
model was served.}
\label{fig:failure_example_6}
\end{figure}

\begin{figure}[p]
\centering
\includegraphics[width=\textwidth]{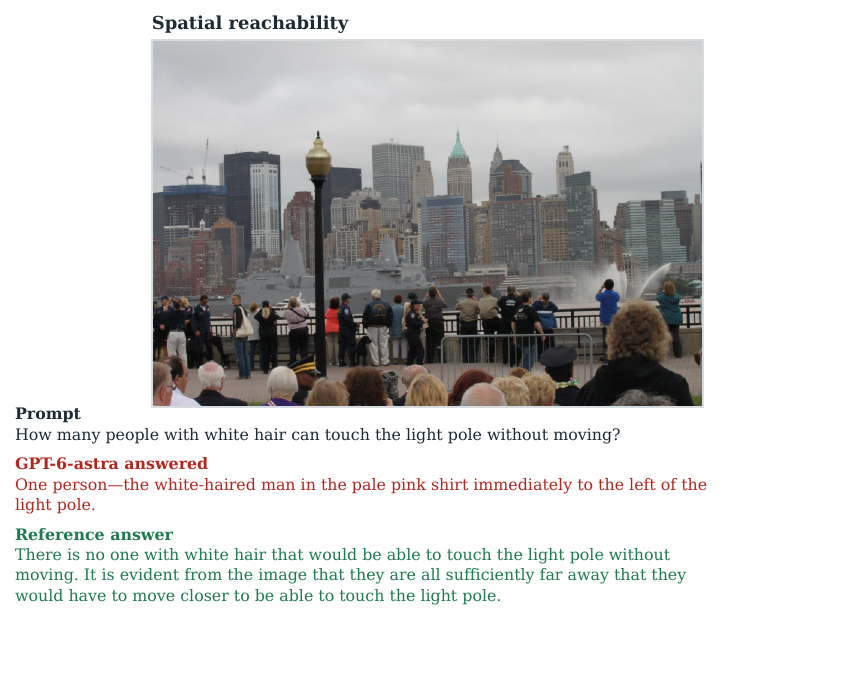}
\caption{\textbf{Failure example: spatial reachability}. The model's answer and the human-written reference answer, with the media the
model was served.}
\label{fig:failure_example_8}
\end{figure}

\begin{figure}[p]
\centering
\includegraphics[width=\textwidth]{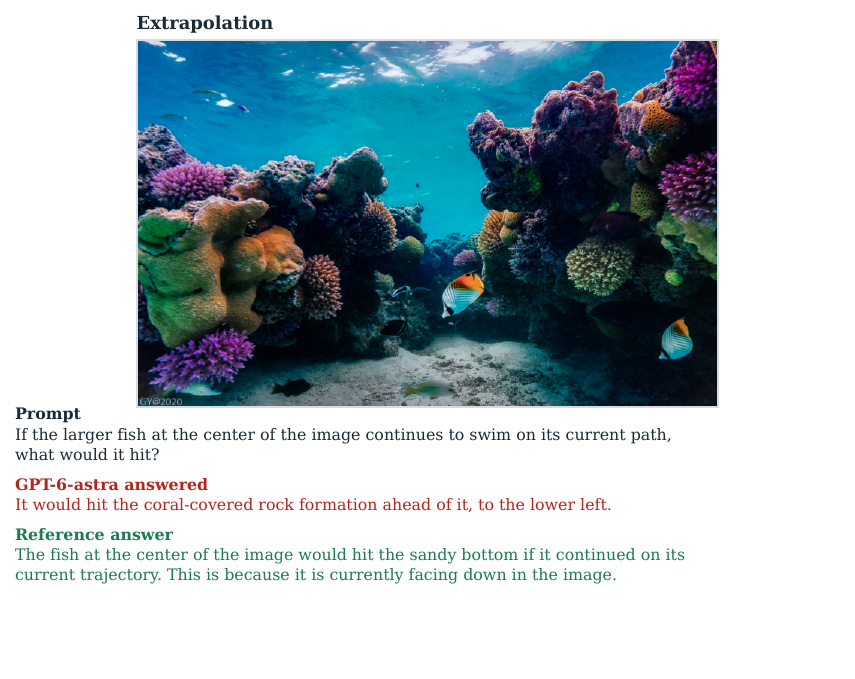}
\caption{\textbf{Failure example: extrapolation}. The model's answer and the human-written reference answer, with the media the
model was served.}
\label{fig:failure_example_10}
\end{figure}

\begin{figure}[p]
\centering
\includegraphics[width=\textwidth]{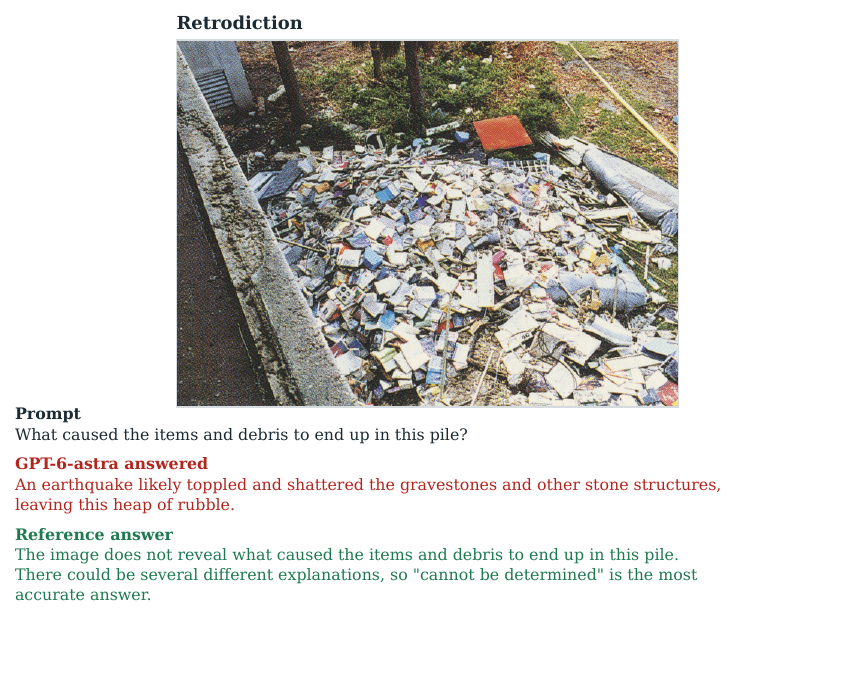}
\caption{\textbf{Failure example: retrodiction}. The model's answer and the human-written reference answer, with the media the
model was served.}
\label{fig:failure_example_12}
\end{figure}

\begin{figure}[p]
\centering
\includegraphics[width=\textwidth]{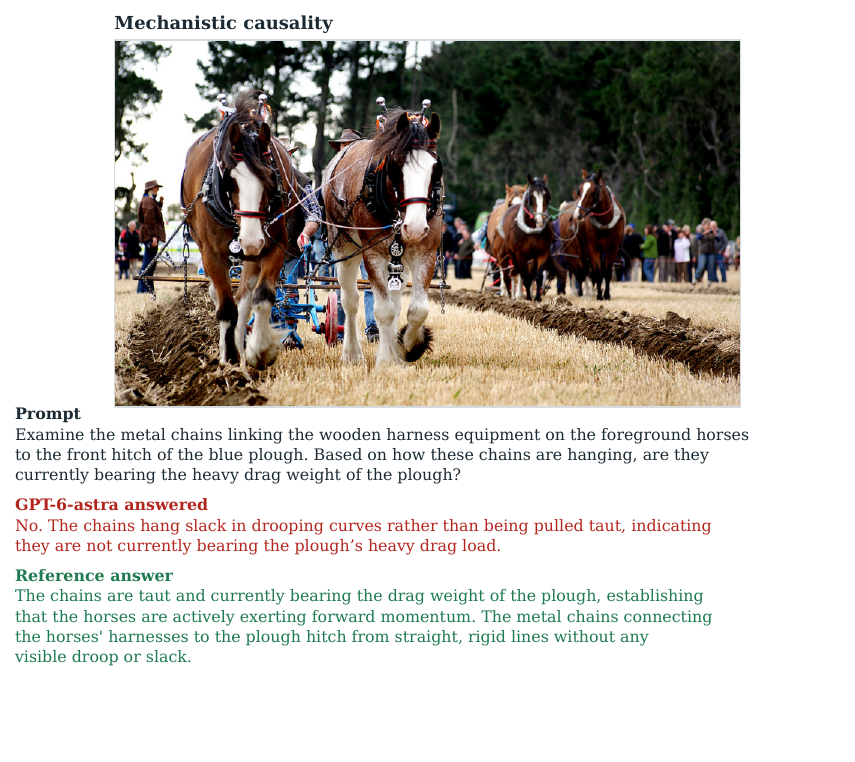}
\caption{\textbf{Failure example: mechanistic causality}. The model's answer and the human-written reference answer, with the media the
model was served.}
\label{fig:failure_example_13}
\end{figure}

\begin{figure}[p]
\centering
\includegraphics[width=\textwidth]{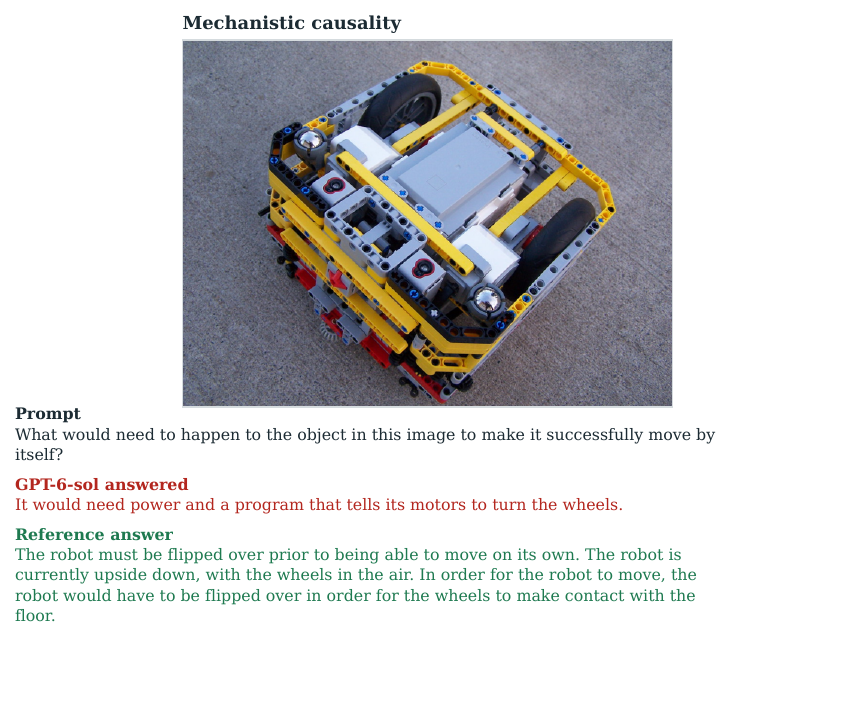}
\caption{\textbf{Failure example: mechanistic causality}. The model's answer and the human-written reference answer, with the media the
model was served.}
\label{fig:failure_example_14}
\end{figure}

\begin{figure}[p]
\centering
\includegraphics[width=\textwidth]{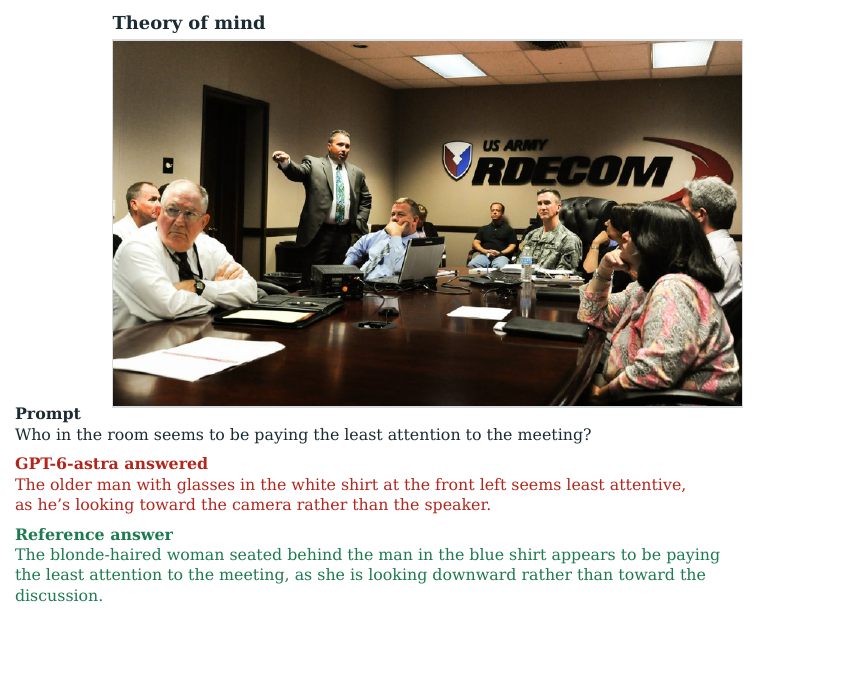}
\caption{\textbf{Failure example: theory of mind}. The model's answer and the human-written reference answer, with the media the
model was served.}
\label{fig:failure_example_15}
\end{figure}

\begin{figure}[p]
\centering
\includegraphics[width=\textwidth]{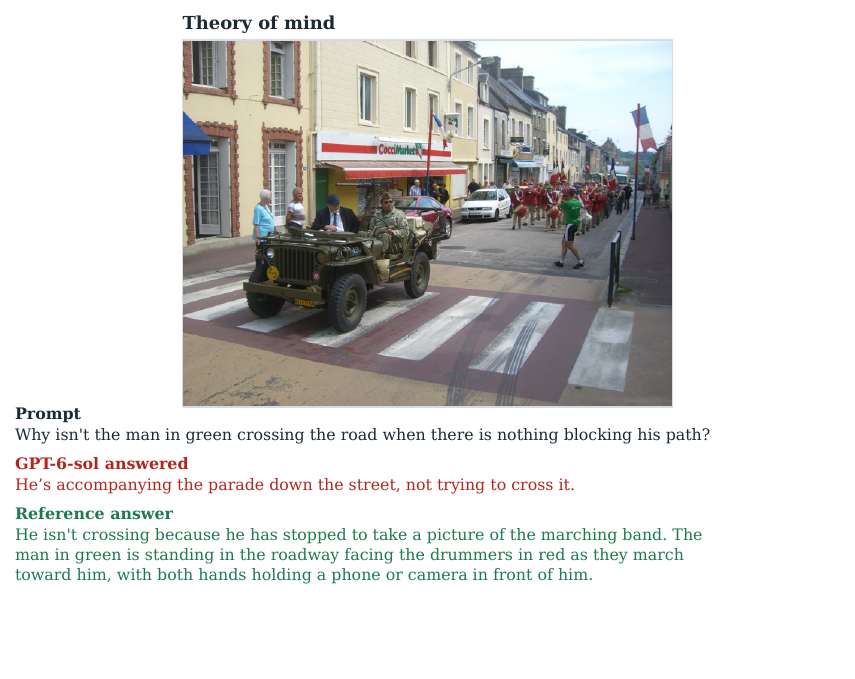}
\caption{\textbf{Failure example: theory of mind}. The model's answer and the human-written reference answer, with the media the
model was served.}
\label{fig:failure_example_16}
\end{figure}

\begin{figure}[p]
\centering
\includegraphics[width=\textwidth]{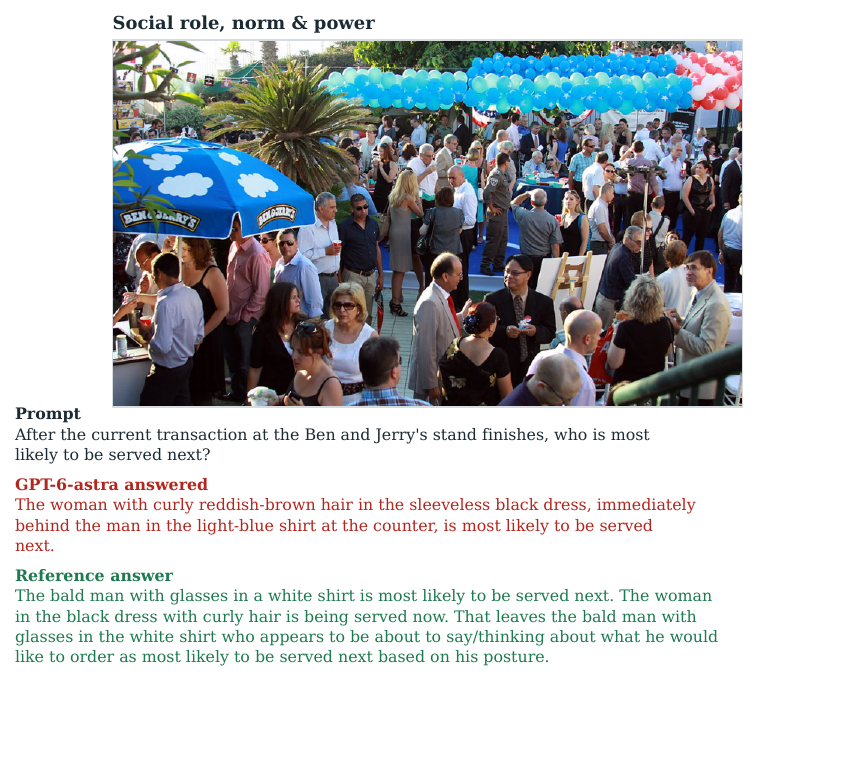}
\caption{\textbf{Failure example: social role, norm \& power}. The model's answer and the human-written reference answer, with the media the
model was served.}
\label{fig:failure_example_18}
\end{figure}

\begin{figure}[p]
\centering
\includegraphics[width=\textwidth]{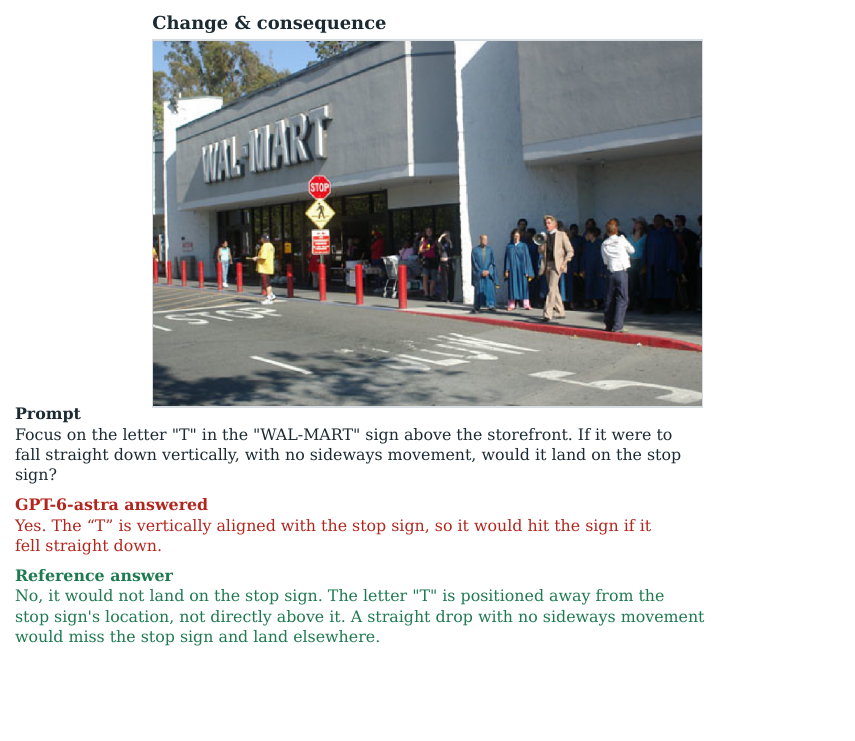}
\caption{\textbf{Failure example: change \& consequence}. The model's answer and the human-written reference answer, with the media the
model was served.}
\label{fig:failure_example_20}
\end{figure}

\begin{figure}[p]
\centering
\includegraphics[width=\textwidth]{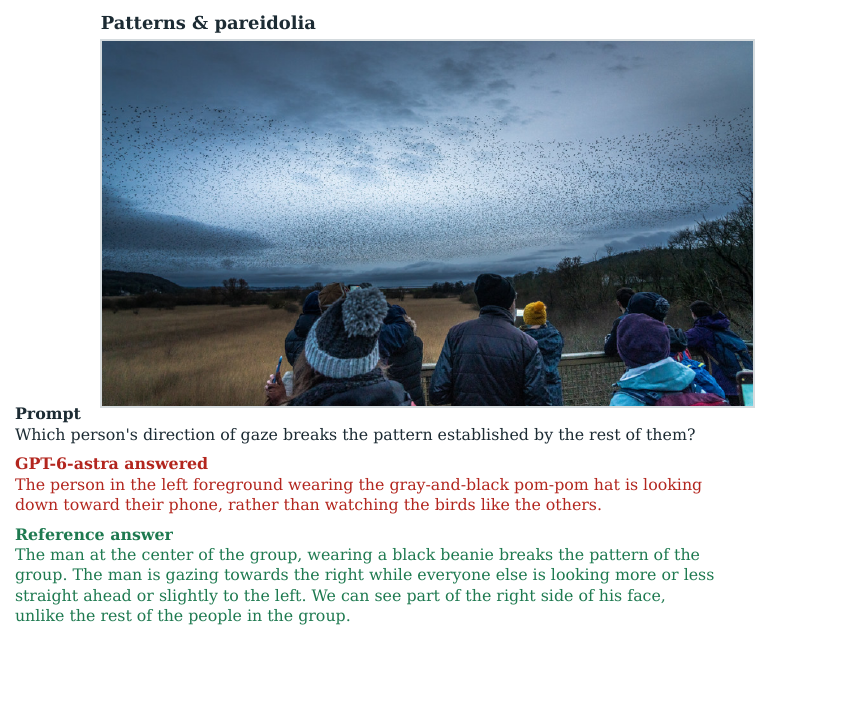}
\caption{\textbf{Failure example: patterns \& pareidolia}. The model's answer and the human-written reference answer, with the media the
model was served.}
\label{fig:failure_example_21}
\end{figure}

\begin{figure}[p]
\centering
\includegraphics[width=\textwidth]{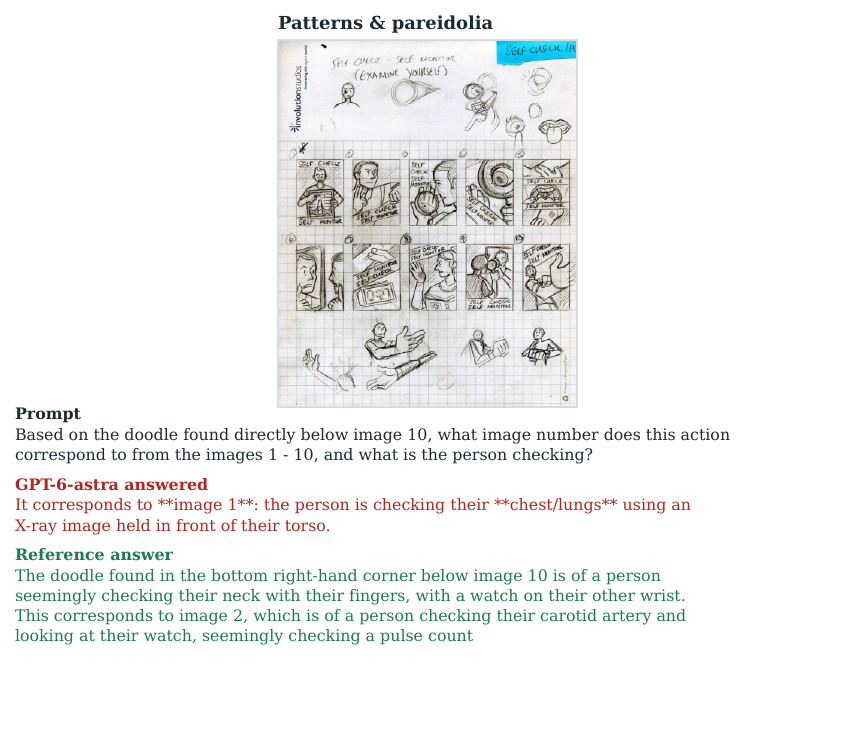}
\caption{\textbf{Failure example: patterns \& pareidolia}. The model's answer and the human-written reference answer, with the media the
model was served.}
\label{fig:failure_example_22}
\end{figure}

This section shows failure examples of HSS (Figure \ref{fig:failure_example_2} to \ref{fig:failure_example_22}): the media the model
was served, the prompt, the model's answer and the human-written reference answer.

% \begin{figure}[p]
% \centering
% \includegraphics[width=\textwidth]{fig_subdomain_case_1.pdf}
% \caption{\textbf{Failure example: hidden \& invisible}. The model's answer and the human-written reference answer, with the media the
% model was served.}
% \label{fig:failure_example_1}
% \end{figure}

% \begin{figure}[p]
% \centering
% \includegraphics[width=\textwidth]{fig_subdomain_case_4.pdf}
% \caption{\textbf{Failure example: spatial reachability}. The model's answer and the human-written reference answer, with the media the
% model was served.}
% \label{fig:failure_example_7}
% \end{figure}

% \begin{figure}[p]
% \centering
% \includegraphics[width=\textwidth]{fig_subdomain_case_5.pdf}
% \caption{\textbf{Failure example: extrapolation}. The model's answer and the human-written reference answer, with the media the
% model was served.}
% \label{fig:failure_example_9}
% \end{figure}

% \begin{figure}[p]
% \centering
% \includegraphics[width=\textwidth]{fig_subdomain_case_6.pdf}
% \caption{\textbf{Failure example: retrodiction}. The model's answer and the human-written reference answer, with the media the
% model was served.}
% \label{fig:failure_example_11}
% \end{figure}

% \begin{figure}[p]
% \centering
% \includegraphics[width=\textwidth]{fig_subdomain_case_9.pdf}
% \caption{\textbf{Failure example: social role, norm \& power}. The model's answer and the human-written reference answer, with the media the
% model was served.}
% \label{fig:failure_example_17}
% \end{figure}

% \begin{figure}[p]
% \centering
% \includegraphics[width=\textwidth]{fig_subdomain_case_10.pdf}
% \caption{\textbf{Failure example: change \& consequence}. The model's answer and the human-written reference answer, with the media the
% model was served.}
% \label{fig:failure_example_19}
% \end{figure}

% ===========================================================================
\section{Agentic Harnesses Analysis}
\label{app:agentic}
% ===========================================================================

Table~\ref{tab:agentic_full} provides all four harness/base pairs behind Table~\ref{tab:agentic}, paired on the same $388$ tasks. All four setups improve the model performance: Claude Code lifts both of its models by $16$--$21$ points at $12$--$14\times$ the base model's thinking tokens, while Codex lifts both of its by $5$--$8$ at $1$--$2\times$ of base model's thinking tokens.

\begin{table}[ht]
\caption{\textbf{All agentic harness/base pairs}, paired on $388$ tasks. \emph{Blank} counts
tasks where the harness returned no answer, scored as zeros; $p$ is two-sided McNemar.}
\label{tab:agentic_full}
\centering
\small
\setlength{\tabcolsep}{4.5pt}
\begin{tabular}{llrrrrrrrr}
\toprule
\textbf{Harness} & \textbf{Base model} & \textbf{Raw} & \textbf{Agent} & \textbf{$\Delta$} &
\textbf{Win} & \textbf{Loss} & \textbf{$p$} & \textbf{Think} & \textbf{Blank} \\
\midrule
Claude Code & Claude-Opus-5    & $30.8$ & $51.3$ & $+20.5$ & $112$ & $30$ & $<10^{-4}$ & $43{,}764$ & $0$ \\
Claude Code & Claude-Fable-5.1 & $41.2$ & $57.7$ & $+16.5$ & $96$ & $30$ & $<10^{-4}$ & $42{,}237$ & $1$ \\
Codex       & GPT-5.6-sol      & $28.4$ & $35.8$ & $+7.5$ & $63$ & $33$ & $0.003$ & $6{,}541$ & $4$ \\
Codex       & GPT-6-astra      & $54.4$ & $59.3$ & $+4.9$ & $53$ & $28$ & $0.007$ & $2{,}040$ & $0$ \\
\bottomrule
\end{tabular}
\end{table}

\paragraph{Trajectories.} Based on agents' trajectories, failures run \emph{longer} than successes: Claude Code takes a median $81$ steps on tasks it gets wrong against $67$ on tasks it gets right around Opus-5 and $65$ against $54$ around Fable-5.1. Codex, which searches least, shows no such gap.

\subsection{Four trajectories}
\label{app:trajectories}

Below, we present four representative agentic runs: two where the agentic workflow turns a single-pass failure into a success (repaired), and two where the reverse occurs, a single-pass success resulting in an agentic failure (broken).

\begin{promptbox}[promptgreen]{Repaired: ``Which team likely won this game of rugby? Why?''}
\small
\textbf{Claude Code + Claude-Opus-5}, social role \& power, video ($30$\,s), $121$ steps.
Single call: \textbf{fail} (all three attempts). Agent: \textbf{pass}.\\[3pt]
\textbf{Reference.} The team in white won: after the whistle several white players raise
both arms above their heads, while the black-and-red side stands with arms down and
shoulders slumped.\\[3pt]
\textbf{Single call.} ``The home side in the dark red-and-black jerseys almost certainly
won'' --- reading a kneeling white player as ``the classic posture of a beaten player''.\\[3pt]
\textbf{Agent.} ``The away team --- the side playing in all white --- almost certainly won.''\\[3pt]
\textbf{Trace.} (3) probes duration and resolution; (4--8) extracts one frame per second
and tiles $36$ of them into a contact sheet; (10--14) opens individual frames hunting the
celebration; (16--18) crops a $240\times110$ region around two players and upscales it to
$1920$\,px; (20) rebuilds the contact sheet at $2$\,fps because $1$\,fps was too coarse;
(24--32) zooms progressively from $900\times450$ down to $420\times220$ on the first
$3.5$ seconds after the whistle; (47) writes a preliminary answer already naming the
white team, then keeps working for $70$ more steps; (50) re-samples the first
$1.6$\,s at $8$\,fps; (63, 86, 106) three attempts to read the stadium scoreboard, all
defeated by LED overexposure; (90) runs scene-cut detection, gets an empty list, and so
establishes that the clip is one continuous shot and really is the final whistle;
(99--104) re-extracts with a contrast boost to trace the goal line and shows the
red-and-black side standing in its own in-goal; (113) ``I have enough evidence to write a
much stronger answer now''; (120) edits its own answer to soften ``the scoreboard reads
LOU'' to ``appears to read'' and deletes a hallucinated advertiser name.\\[3pt]
\textbf{Why it worked.} The cue was in the frame but brief. Re-sampling the same three
seconds at $1$, $2$ and $8$\,fps and zooming five-fold on the celebrating cluster turned
a glance into a measurement, and the pitch-line and scene-cut checks supplied
corroboration the single call never sought.
\end{promptbox}

\begin{promptbox}[promptgrey]{Broken: ``If the lightbulb hanging above the bald-headed man with red glasses were to fall, where would it end up relative to his body?''}
\small
\textbf{Claude Code + Claude-Opus-5}, change \& consequence, image, $36$ steps.
Single call: \textbf{pass} (all three attempts). Agent: \textbf{fail}.\\[3pt]
\textbf{Reference.} Behind him. The bulb is not lighting his face --- no shadow under the
chin, only a sheen on the top of his head --- so it hangs behind his head from the
camera's viewpoint.\\[3pt]
\textbf{Single call.} ``The fixture hangs from the ceiling well behind him\ldots it would
land just behind his right shoulder rather than squarely on it.''\\[3pt]
\textbf{Agent.} ``It would land on his head --- specifically on the front-left of his
shaved head\ldots It would \emph{not} land behind him.''\\[3pt]
\textbf{Trace.} (3--4) reads the image and its dimensions; (8) writes a preliminary
answer; (10--19) crops the bulb at $300\%$, the head at $400\%$ and the glasses at
$500\%$ and reads each back; (24) dumps raw grayscale pixel rows across the support pole
at five heights to test whether the camera is rolled; (26) does the same across the bulb
and computes an intensity-weighted centroid at $x \approx 449$, then scans rows
$y = 150$--$240$ for the left edge of the head; (28) measures the skull's apex and right
edge; (30) ``I have precise measurements now''; (31) draws a red plumb line at $x = 449$
down the whole image and crops it for inspection; (34) ``The overlay confirms the plumb
line''; (35) writes the final answer --- ``I can just read off what that line passes
through'' --- and adds a section titled \emph{One caveat about depth}: ``If it hangs, say,
half a metre behind him, it would fall past the back of his head'', which it then
overrules on the grounds that his forehead ``is clearly lit from above-front-left by this
light''.\\[3pt]
\textbf{Why it broke.} The tools answered the wrong question precisely. A 2D overlap was
converted into a 3D plumb line to sub-pixel accuracy, and the arithmetic gave the agent
enough confidence to overrule the depth objection it had itself raised --- and to
misread the very lighting evidence that settles the task. This is the failure mode of
Table~\ref{tab:failure_shift} in miniature: reading a 2D overlap as 3D alignment is the
one cause the agent loop does not reduce.
\end{promptbox}

\begin{promptbox}[promptgreen]{Repaired: ``\ldots he approaches a couple. What does he notice that causes him to approach them?''}
\small
\textbf{Codex + GPT-6-astra}, theory of mind, video ($20$\,min at $320\times240$),
$21$ steps. Single call: \textbf{fail}. Agent: \textbf{pass}.\\[3pt]
\textbf{Reference.} He notices the woman drop her handkerchief, and approaches to
retrieve it and return it.\\[3pt]
\textbf{Single call.} ``He notices the woman peering at him through her opera glasses.''\\[3pt]
\textbf{Agent.} ``He notices that the woman\ldots has dropped her handkerchief over the
edge of their theater box.''\\[3pt]
\textbf{Trace.} (5) probes the file; (6) builds one $5\times6$ contact sheet covering the
first $300$\,s at one frame per $10$\,s, with the timestamp burned into each tile;
(7) reads the sheet and locates the two-chair act near $t \approx 205$\,s; (8--9) a
re-extraction fails and the image read errors out; (10) debugs its own \texttt{ffmpeg}
invocation, moving the duration flag and switching to an absolute burned-in clock;
(12--14) zooms $t = 215$--$223$ at $2$\,fps and takes one $960\times720$ close-up at
$t = 250.8$\,s; (15--16) ``He notices the woman drop her handkerchief'' and writes the
answer; (17--21) a verification pass over $t = 216$--$246$ at $2$\,fps confirms it.\\[3pt]
\textbf{Why it worked.} Timestamped contact sheets localised a twenty-minute film to a
ten-second window, and a three-fold upscale of a $320\times240$ source resolved an object
a few pixels wide --- the detail the single call replaced with a plausible story.
\end{promptbox}

\begin{promptbox}[promptgrey]{Broken: ``For what reason does the woman with darker sleeves and a dark apron take objects from the other workers?''}
\small
\textbf{Codex + GPT-6-astra}, theory of mind, video ($148$\,s), $27$ steps.
Single call: \textbf{pass}. Agent: \textbf{fail} (rubric $0.5$: the job title is right,
the evidence is not).\\[3pt]
\textbf{Reference.} A quality inspection: she takes each piece, examines it closely, and
\emph{returns it} to the worker.\\[3pt]
\textbf{Single call.} ``She takes them briefly, holds them up and turns them to examine
them, then hands them back and moves to another worker.''\\[3pt]
\textbf{Agent.} ``She is acting as an inspector or supervisor, checking that the pieces
have been made correctly'' --- describing her as taking a piece and moving on, never
noting the hand-back.\\[3pt]
\textbf{Trace.} (6--8) overview sheet at one frame per $10$\,s; (9--11) re-samples
$t = 8$--$63$ at one frame per $3$\,s plus a full-resolution frame; (12--15) two more
sheets covering the hand-off window and the tail of the clip; (16) writes a first answer
--- ``inspecting and gathering their output'' --- and in the same command builds a
$40$-tile sheet of $t = 25$--$45$ at $2$\,fps cropped to her work area; (19--22) crops
the hand-off at $1$\,fps; (23--25) the densest look of the run, $t = 52$--$55$ at
$4$\,fps upscaled two-fold; (26) revises the answer, changing ``inspecting and gathering
their output'' to ``checking that the pieces have been made correctly''.\\[3pt]
\textbf{Why it broke.} It sampled the right seconds at the right rate and still narrated
only half the exchange. Its one revision tightened the job title rather than re-examining
the action, so the return --- the evidence the rubric asks for --- never entered the
answer. More looking does not help when the agent has already decided what it is looking
at.
\end{promptbox}

% ===========================================================================
\section{Reasoning Effort in Full}
\label{app:effort}
% ===========================================================================

Table~\ref{tab:effort_full} gives per-subdomain accuracy at every reasoning level, the
numbers behind Figure~\ref{fig:effort_by_domain}. At the domain level, moving from the
lowest to the highest setting gives GPT-6-astra $+8.2$ (temporal), $+12.5$ (physical),
$+15.5$ (social) and $+7.8$ (abstract); Claude-Opus-5.5 $+6.7$, $+9.1$, $+0.9$ and
$+4.9$; and Gemini-3.8-flash $+0.7$, $+5.7$, $+3.6$ and $-1.0$. Seven of the $33$
model $\times$ subdomain cells end no better than they started, and three of
Gemini-3.8-flash's eleven end lower. In only five of the twelve model $\times$ domain
curves is the best score at the highest available setting.

\begin{table}[h]
\caption{\textbf{Accuracy by subdomain and reasoning effort} (\%), all $522$ tasks.
Columns are each deployment's levels in order --- five for GPT-6-astra
(\texttt{low}--\texttt{max}), four for Claude-Opus-5.5, three for Gemini-3.8-flash ---
and $\Delta$ is last minus first.}
\label{tab:effort_full}
\centering
\small
\setlength{\tabcolsep}{5pt}
\begin{tabular}{lrrrrrr}
\toprule
\textbf{Subdomain} & \multicolumn{5}{c}{\textbf{Reasoning level, lowest $\rightarrow$ highest}} & \textbf{$\Delta$} \\
\midrule
\multicolumn{7}{l}{\textit{GPT-6-astra}}\\
\quad Hidden \& Invisible & $48$ & $50$ & $52$ & $57$ & $60$ & $+12$ \\
\quad Affordance \& Feasibility & $39$ & $41$ & $44$ & $49$ & $56$ & $+17$ \\
\quad Spatial ``Alien Viewpoint'' & $28$ & $30$ & $40$ & $46$ & $44$ & $+16$ \\
\quad Spatial Reachability & $22$ & $22$ & $26$ & $22$ & $22$ & $+0$ \\
\quad Extrapolation & $41$ & $45$ & $47$ & $49$ & $53$ & $+12$ \\
\quad Retrodiction & $50$ & $48$ & $64$ & $50$ & $56$ & $+6$ \\
\quad Mechanistic Causality & $55$ & $58$ & $64$ & $61$ & $61$ & $+6$ \\
\quad Theory of Mind & $29$ & $31$ & $37$ & $39$ & $45$ & $+16$ \\
\quad Social Role, Norm \& Power Dynamics & $32$ & $31$ & $46$ & $42$ & $47$ & $+15$ \\
\quad Change \& Consequence & $39$ & $40$ & $43$ & $51$ & $47$ & $+9$ \\
\quad Patterns \& Pareidolia & $44$ & $44$ & $44$ & $50$ & $50$ & $+6$ \\
\addlinespace
\multicolumn{6}{l}{\textit{Claude-Opus-5.5}}\\
\quad Hidden \& Invisible & $24$ & $34$ & $41$ & $40$ &  & $+16$ \\
\quad Affordance \& Feasibility & $41$ & $54$ & $41$ & $56$ &  & $+15$ \\
\quad Spatial ``Alien Viewpoint'' & $38$ & $46$ & $44$ & $40$ &  & $+2$ \\
\quad Spatial Reachability & $59$ & $63$ & $56$ & $59$ &  & $+0$ \\
\quad Extrapolation & $35$ & $39$ & $43$ & $43$ &  & $+8$ \\
\quad Retrodiction & $40$ & $40$ & $50$ & $48$ &  & $+8$ \\
\quad Mechanistic Causality & $42$ & $42$ & $45$ & $45$ &  & $+3$ \\
\quad Theory of Mind & $25$ & $27$ & $25$ & $27$ &  & $+2$ \\
\quad Social Role, Norm \& Power Dynamics & $41$ & $46$ & $41$ & $41$ &  & $+0$ \\
\quad Change \& Consequence & $51$ & $47$ & $60$ & $59$ &  & $+7$ \\
\quad Patterns \& Pareidolia & $38$ & $44$ & $41$ & $38$ &  & $+0$ \\
\addlinespace
\multicolumn{5}{l}{\textit{Gemini-3.8-flash}}\\
\quad Hidden \& Invisible & $33$ & $34$ & $40$ &  &  & $+7$ \\
\quad Affordance \& Feasibility & $32$ & $34$ & $34$ &  &  & $+2$ \\
\quad Spatial ``Alien Viewpoint'' & $34$ & $48$ & $38$ &  &  & $+4$ \\
\quad Spatial Reachability & $41$ & $41$ & $52$ &  &  & $+11$ \\
\quad Extrapolation & $33$ & $31$ & $39$ &  &  & $+6$ \\
\quad Retrodiction & $44$ & $44$ & $42$ &  &  & $-2$ \\
\quad Mechanistic Causality & $42$ & $39$ & $39$ &  &  & $-3$ \\
\quad Theory of Mind & $33$ & $37$ & $35$ &  &  & $+2$ \\
\quad Social Role, Norm \& Power Dynamics & $32$ & $39$ & $37$ &  &  & $+5$ \\
\quad Change \& Consequence & $49$ & $50$ & $50$ &  &  & $+1$ \\
\quad Patterns \& Pareidolia & $44$ & $41$ & $38$ &  &  & $-6$ \\
\bottomrule
\end{tabular}
\end{table}

% ===========================================================================
\section{Video Frame Budget in Full}
\label{app:frames}
% ===========================================================================

Table~\ref{tab:frames} gives the headline video-frame sweep, which runs over the $234$ video tasks of HSS; frames are sampled at $2$~fps and thinned uniformly to the cap. Every budget spans the whole clip and the comparison is one of temporal \emph{resolution} rather than coverage. The $500$-frame column is the configuration used everywhere else in the paper, scored here on the first attempt alone, which is why it
sits a few tenths from the three-attempt video column of Table~\ref{tab:main}.

\begin{table}[ht]
\caption{\textbf{Video frame budget.} Pass@1 (\%) on the $234$ video tasks at each
per-request frame cap, sampled at $2$~fps and thinned uniformly to fit. $\Delta$ is
$500$ minus $10$ frames; \emph{best} is the budget at which each model peaks.}
\label{tab:frames}
\centering
\small
\begin{tabular}{lrrrrrrr}
\toprule
\textbf{Model} & \textbf{10} & \textbf{50} & \textbf{100} & \textbf{200} & \textbf{500} &
\textbf{$\Delta$} & \textbf{best} \\
\midrule
GPT-6-astra     & $33.8$ & $41.9$ & $45.3$ & $47.0$ & $\mathbf{51.7}$ & $\mathbf{+17.9}$ & $500$ \\
Claude-Opus-5.5 & $30.3$ & $32.1$ & $32.5$ & $\mathbf{37.2}$ & $36.3$ & $+6.0$ & $200$ \\
GPT-6-sol       & $23.1$ & $27.8$ & $29.1$ & $26.1$ & $\mathbf{31.2}$ & $+8.1$ & $500$ \\
Claude-Opus-5   & $18.8$ & $19.7$ & $\mathbf{28.6}$ & $27.4$ & $27.4$ & $+8.5$ & $100$ \\
\bottomrule
\end{tabular}
\end{table}

Three things are worth separating. First, densen sampling helps. Every model improves from $10$ to $500$ frames. Second, it saturates. The $10 \rightarrow 100$ step is
worth $+11.5$, $+6.0$, $+2.1$ and $+9.8$ points to the four models; the $100 \rightarrow 500$ step, a five-fold further increase in tokens, is worth $+6.4$, $+2.1$, $+3.8$ and $-1.3$. Third, at this sample size most of it is not resolved: the per-cell interval is about $\pm 6$ points and only GPT-6-astra's end-to-end gain has non-overlapping intervals at its endpoints. We therefore read the table as ruling out a large monotone gain from temporal resolution beyond roughly $100$ frames, and as unable to resolve a small one.

% ===========================================================================
\section{Judge Alignment in Full}
\label{app:judge}
% ===========================================================================

This section expands the summary in Section~\ref{sec:judge_alignment}. Five models spanning the leaderboard, including GPT-6-astra, Gemini-3.8-flash, Claude-Fable-5, Claude-Opus-5 and GLM-5.3-flash were re-graded end to end by three judges from three vendors, on identical rubrics, reference answers and grading prompts. Per-model scores under each judge are in Table~\ref{tab:judge_alignment}.

\begin{table}[h]
\caption{\textbf{Cross-judge alignment.} Pass@1 (\%) for the same five models under three judges from three vendors, single attempt per task. The rightmost column is the largest
disagreement between any two judges about a model. Every pairwise rank correlation is $\rho = 1.0$: the three judges
disagree about levels, never about order. Agreement statistics for the same runs are plotted in Figure~\ref{fig:judge_alignment}.}
\label{tab:judge_alignment}
\centering
\small
\begin{tabular}{lrrrr}
\toprule
 & \multicolumn{3}{c}{\textbf{Judge}} & \\
\cmidrule(lr){2-4}
\textbf{Model} & \textbf{Claude-Opus-5} & \textbf{GPT-5.6-terra} & \textbf{Gemini-3.6-flash} & \textbf{Spread} \\
\midrule
GPT-6-astra (max)       & $53.8$ & $50.8$ & ${54.2}$ & $3.4$ \\
Gemini-3.8-flash (high) & $42.7$ & $40.8$ & ${43.5}$ & $2.7$ \\
Claude-Fable-5 (max)    & ${33.8}$ & $30.2$ & $33.0$ & $3.6$ \\
Claude-Opus-5 (max)     & $31.5$ & $30.0$ & ${31.7}$ & $1.7$ \\
GLM-5.3-flash (max)     & $26.9$ & $25.2$ & ${27.1}$ & $1.9$ \\
\midrule
Mean over the five      & $37.7$ & $35.4$ & $37.9$ & \\
\bottomrule
\end{tabular}
\end{table}

\paragraph{The judges agree, and where they differ they differ by a constant.} Pairwise agreement is $95.8$--$97.2\%$ at the criterion level and $96.2$--$97.5\%$ at the item level, with Cohen's $\kappa$ between $0.914$ and $0.946$ on both units (Figure~\ref{fig:judge_alignment}b); all three judges return the same verdict on $95.2\%$ of items. The disagreement that remains is almost entirely a level shift rather than a reordering. GPT-5.6-terra is the strict judge, scoring every one of the five models on average $2.4$ points below the mean of the other two, while Claude-Opus-5 and Gemini-3.6-flash sit within $0.3$ points of each other on average. Spearman rank correlation between judges is exactly $1.0$ for all three pairs, so no conclusion in this paper that rests on model ordering changes under a different grader.

\paragraph{Self-preference is not detectable at this scale.}
LLM judges are known to favour their own family's outputs
\citep{panickssery2024selfpreference}, which matters here because our primary judge shares a vendor with multiple evaluated models. To evaluate potential vendor bias, we first calculated each judge's overall leniency—defined as the difference between the score it gives a model and the average score given by the other two judges:
\begin{itemize}
    \item Claude-Opus-5: +1.1 points,
    \item GPT-5.6-terra: -2.4 points,
    \item Gemini-3.6-flash: +1.3 points.
\end{itemize}
Subtracting these baseline leniencies isolates same-vendor interaction effects (residuals):
\begin{itemize}
    \item GPT-5.6-terra judging GPT-6-astra: -0.80 points, 
    \item Gemini-3.6-flash judging Gemini-3.8-flash: +0.38 points,
    \item Claude-Opus-5 judging Claude-Fable-5: +1.09 points,
    \item Claude-Opus-5 judging itself: -0.44 points.
\end{itemize}
All four residuals are small and two have negative signs (contradicting the expectation of self-preference). We interpret this not as proof that vendor bias is entirely absent, but as finding no detectable evidence of it at the resolution HSS currently supports.

% ===========================================================================
\end{document}